\documentclass[sigconf]{acmart}

\usepackage{CJKutf8}
\usepackage{booktabs}
\usepackage{graphicx}
\usepackage{array}
\usepackage{bm}
\usepackage{amsmath}
\renewcommand{\abstractname}{\MakeUppercase{Abstract}}
\renewcommand{\refname}{\MakeUppercase{References}}
\usepackage{titlesec}

\titleformat{\section}[block]{\normalfont\Large\bfseries}{\thesection}{1em}{\MakeUppercase}

\usepackage{balance}
\usepackage{threeparttable}
\usepackage{multirow}
\usepackage{makecell}

\usepackage{latexsym}
\newcommand{\tabincell}[2]{\begin{tabular}{@{}#1@{}}#2\end{tabular}}
\usepackage[T1]{fontenc}

\usepackage[utf8]{inputenc}
\usepackage{float}

\usepackage{microtype}

\usepackage{xcolor}

\newcommand{\eat}[1]{{}}

\def\BibTeX{{\rm B\kern-.05em{\sc i\kern-.025em b}\kern-.08em
    T\kern-.1667em\lower.7ex\hbox{E}\kern-.125emX}}
\usepackage{cite}
\usepackage{amsmath,amssymb,amsfonts}
\usepackage{algorithmic}
\usepackage[ruled,linesnumbered,noline,noend]{algorithm2e}

\SetCommentSty{mycommfont}

\SetNlSty{}{}{}

\let\oldnl\nl
\newcommand\nonl{%
  \renewcommand{\nl}{\let\nl\oldnl}}
  
\usepackage{hyperref}
\hypersetup{
    colorlinks=true,
    linkcolor=blue,}
\AtBeginDocument{%
  \providecommand\BibTeX{{%
    \normalfont B\kern-0.5em{\scshape i\kern-0.25em b}\kern-0.8em\TeX}}}

\copyrightyear{2023}
\acmYear{2023}
\setcopyright{acmlicensed}\acmConference[CIKM '23]{Proceedings of the 32nd ACM International Conference on Information and Knowledge Management}{October 21--25, 2023}{Birmingham, United Kingdom}
\acmBooktitle{Proceedings of the 32nd ACM International Conference on Information and Knowledge Management (CIKM '23), October 21--25, 2023, Birmingham, United Kingdom}
\acmPrice{15.00}
\acmDOI{10.1145/3583780.3614905}
\acmISBN{979-8-4007-0124-5/23/10}

\begin{document}

\title{Hallucination Detection: Robustly Discerning Reliable Answers in Large Language Models}

\author{Yuyan Chen}
\authornote{Work done while this author was an intern at Microsoft Research.}
\email{chenyuyan21@m.fudan.edu.cn}
\orcid{0000-0002-4381-486X}
\affiliation{%
  \institution{Shanghai Key Laboratory of Data Science, School of Computer Science, Fudan University}
  \city{Shanghai}
  \country{China}
}

\author{Qiang Fu}
\authornote{The corresponding authors.}
\email{qifu@microsoft.com}
\orcid{0000-0002-5821-7267}
\affiliation{%
  \institution{Microsoft}
  \city{Beijing}
  \country{China}
}

\author{Yichen Yuan}
\email{axclbkj@gmail.com}
\orcid{0009-0009-1321-3850}
\affiliation{%
  \institution{Shanghai Key Laboratory of Data Science}
  \city{Shanghai}
  \country{China}
}

\author{Zhihao Wen}
\email{zhwen.2019@phdcs.smu.edu.sg}
\orcid{0000-0002-7688-5381}
\affiliation{%
  \institution{Singapore Management University}
  \city{Singapore}
  \country{Singapore}
}

\author{Ge Fan}
\email{ge.fan@outlook.com}
\orcid{0000-0001-5653-1626}
\affiliation{%
  \institution{Tencent}
  \city{Shenzhen}
  \country{China}
}

\author{Dayiheng Liu}
\email{liudayiheng.ldyh@alibaba-inc.com}
\orcid{0000-0001-5653-1626}
\affiliation{%
  \institution{DAMO Academy}
  \city{Hangzhou}
  \country{China}
}

\author{Dongmei Zhang}
\email{dongmeiz@microsoft.com}
\orcid{0000-0002-9230-2799}
\affiliation{%
  \institution{Microsoft}
  \city{Beijing}
  \country{China}
}

\author{Zhixu Li}
\authornotemark[2]
\email{zhixuli@fudan.edu.cn}
\orcid{0000-0003-2355-288X}
\affiliation{%
  \institution{Shanghai Key Laboratory of Data Science, School of Computer Science, Fudan University}
  \city{Shanghai}
  \country{China}
}

\author{Yanghua Xiao}
\authornotemark[2]
\email{shawyh@fudan.edu.cn}
\orcid{0000-0001-8403-9591}
\affiliation{%
  \institution{Shanghai Key Laboratory of Data Science, School of Computer Science, Fudan University, Fudan-Aishu Cognitive Intelligence Joint Research Center}
  \city{Shanghai}
  \country{China}
}


\renewcommand{\shortauthors}{Yuyan Chen et al.}

\begin{abstract}
Large Language Models (LLMs) have gained widespread adoption in various natural language processing tasks, including question answering and dialogue systems. However, a major drawback of LLMs is the issue of hallucination, where they generate unfaithful or inconsistent content that deviates from the input source, leading to severe consequences.
In this paper, we propose a robust discriminator named RelD to effectively detect hallucination in LLMs' generated answers. RelD is trained on the constructed RelQA, a bilingual question-answering dialogue dataset along with answers generated by LLMs and a comprehensive set of metrics. 
Our experimental results demonstrate that the proposed RelD successfully detects hallucination in the answers generated by diverse LLMs. Moreover, it performs well in distinguishing hallucination in LLMs' generated answers from both in-distribution and out-of-distribution datasets. Additionally, we also conduct a thorough analysis of the types of hallucinations that occur and present valuable insights.
This research significantly contributes to the detection of reliable answers generated by LLMs and holds noteworthy implications for mitigating hallucination in the future work.
\end{abstract}



\maketitle

\section{INTRODUCTION}
\begin{figure}[t]
  \centering
  \includegraphics[width=0.88\linewidth]{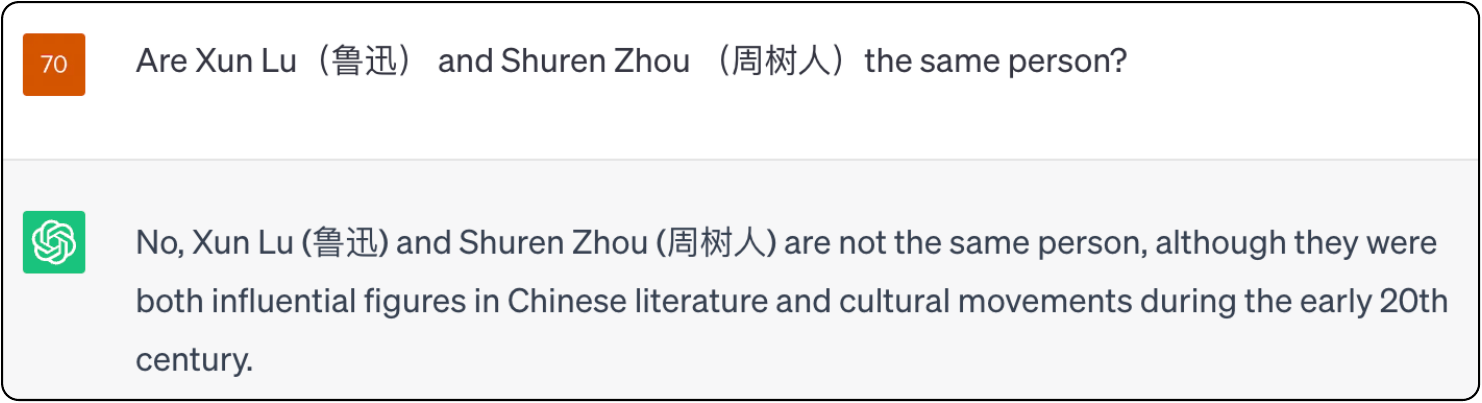}
  \caption{\footnotesize The given answer, produced by ChatGPT, exhibits ``Hallucinations'' by incorrectly treating ``Shuren Zhou'' and ``Xun Lu'' as separate individuals, despite they referring to the same person.}
  \label{fig:example}
  \vspace{-1em}
\end{figure}

Large language models (LLMs) have revolutionized various fields~\citep{zhao2023survey,chen2023hadamard,li2024kenet}, including logical reasoning~\citep{bang2023multitask,liu2023evaluating}, question answering~\citep{chen2024talk,chen2023xmqas,yang2021amqan}, text generation~\citep{zong2024proswitch,chen2023mapo,chen2022grow}, and vertical domains~\citep{liu2023summary,chen2024temporalmed,xia2024aicodereval}. However, LLMs encounter numerous challenges that hinder their optimal performance. These challenges include the inability to update knowledge in real-time~\citep{chenapreliminary}, the lack of genuine emotion and thought~\citep{chalmers2023could,chen2023can}, and the generation of long-winded and verbose answers~\citep{jiang2023graphologue}, among others. Notably, one of the most critical failures is the presence of factual errors in the generated text~\citep{borji2023categorical}, which gives rise to ``Hallucinations'' as depicted in Fig~\ref{fig:example}.
The existence of such ``Hallucinations'' poses a severe hindrance to the widespread adoption of LLMs in non-chatbot scenarios, particularly in domains like medicine and finance where factual accuracy is crucial. The potential risks associated with erroneous information can lead to significant economic losses or even jeopardize human safety~\citep{aiyappa2023can}. Consequently, the elimination of factual errors in LLMs has become an essential requirement in both industry and academia.

The issue of hallucinations in natural text generation has long been acknowledged by researchers~\citep{ji2023survey, li2023helma,li2023evaluating}, and the causes of these hallucinations are complex and multifaceted. On one hand, the large-scale data corpus employed for training LLMs unavoidably contains some erroneous information, which gets learned and stored in the model parameters~\citep{madotto2020language,petroni2019language,roberts2020much}. Consequently, when generating text, LLMs tend to prioritize their parameterized knowledge, thereby resulting in the production of hallucinatory content~\citep{longpre2021entity}.
On the other hand, the decoder component of LLMs is typically trained using maximum likelihood estimation~\citep{bengio2015scheduled,ranzato2015sequence}. During training, ground-truth serves as the input prefix for predicting subsequent tokens. However, during inference, the next token is predicted based on the generated history sequence~\citep{he2019exposure}. This discrepancy in the prediction process makes it easier for hallucinations to occur.

Existing research on detecting hallucinations of LLMs' generated answers primarily encompasses statistical, model-based, and human-based evaluations~\citep{ji2023survey,li2023halueval}.
Statistical evaluation involves direct calculation of vocabulary matching between the generated text and reference target text, employing metrics such as ROUGE~\citep{lin2004rouge} and BLEU~\citep{papineni2002bleu}. Some studies also utilize the Knowledge F1 (KF1)~\citep{shuster2021retrieval} metric to reduce knowledge
hallucination in state-of-the-art chatbots. This KF1 metric is particularly suitable for detecting hallucinations in knowledge dialogue scenarios. Additionally, Shen et al.~\citep{shen2023chatgpt} conduct a large-scale assessment, including correctness and unanswerable question identification, to evaluate ChatGPT's reliability in generic question-answering scenarios. Ye et al.~\citep{ye2023assessing} undertake a preliminary study to assess the robustness, consistency, and credibility of LLM systems. However, these metrics rely on vocabulary matching and surface-level metrics, which may not capture semantic coherence or accurately detect hallucinations.  
Model-based evaluation defines the hallucination score based on the entailment probability between the source text and the generated text. This involves judging whether a hypothesis (i.e., generated text) is entailed by the premise (i.e., reference text). Model-based evaluation incorporates various metrics, including Information Extraction (IE)-based metrics, QA-based metrics~\citep{durmus2020feqa,scialom2021questeval,wang2020asking}, Natural Language Inference (NLI) metrics~\citep{dziri2021evaluating, falke2019ranking,honovich2021q}, Faithfulness Classification metrics~\citep{liu2021token,honovich2021q,zhou2020detecting}, and LM-based metrics~\citep{filippova2020controlled,tian2019sticking}. For example, Honovich et al.\citep{honovich2021q} employ the Q$^2$ method of QA systems to assess the consistency between the response and external knowledge. Azaria et al.\citep{azaria2023internal} utilize the internal state and hidden layer activations of LLMs to detect the truthfulness of generated statements.
However, these methods lack a comprehensive set of metrics to effectively balance the advantages and disadvantages of different evaluation criteria. As a result, models often rely heavily on single labels without considering a broader range of factors.
Human-based evaluation involves scoring hallucinatory text or directly comparing it with the ground truth~\citep{santhanam2021rome,shuster2021retrieval}, which inevitably increases research costs.

To address these limitations and achieve a more balanced approach, we combine automatic metrics with model-based evaluation, which aims to align with trends observed in human evaluation scores~\citep{lee2022factuality}. 
Therefore, in this work, we focus on building a robust discriminator, RelD, which is trained on the constructed RelQA, a bilingual question-answering dialogue dataset along with answers generated by LLMs and a comprehensive set of metrics, in order to effectively detect hallucinations in the generated answers of LLMs. Specifically, 
the RelQA dataset comprises 274,426 samples, encompassing diverse sources such as Wikipedia, Baidu Zhidao, Bing user queries, and Chinese high school reading comprehension, etc. These datasets cover a range of domains including Wikipedia, news, education, and stories, utilizing various formats such as extractive reading comprehension and multiple-choice questions. To comprehensively evaluate LLMs' generated answers in the RelQA dataset, we adopt a set of comprehensive metrics, including LLM-assessment metrics, human metrics, machine metrics, and composite metrics. 
Additionally, we introduce a novel and robust discriminator, RelD, which is trained on RelQA, to detect hallucinations and analyze the types of them present in the generated answers of LLMs. Our experimental results demonstrate that RelD performs admirably in detecting hallucinations across diverse LLMs and for both in-distribution and out-of-distribution datasets.
Our contributions in this paper can be outlined as follows:
\begin{itemize}
\setlength{\itemsep}{0pt}
    \item We design a novel and robust discriminator RelD, which aims to detect hallucinations in the generated answers of various LLMs. 
    \item In order to train RelD, we construct RelQA, a bilingual question-answering dialogue dataset along with answers generated by LLMs and a comprehensive set of metrics, including LLM-assessment metrics, human metrics, machine metrics, and composite metrics.
    \item Our experimental results demonstrate that the discriminator RelD effectively detects hallucinations in the answers generated by different LLMs, exhibiting proficiency in both in-distribution and out-of-distribution datasets. Additionally, we make detailed analysis for types of hallucinations and provide valuable insights into the underlying causes of hallucination.
\end{itemize}

\section{DATA CONSTRUCTION}
In this section, we present the process of constructing RelQA. We begin by using questions from various existing nine datasets as inputs to different LLMs to generate corresponding answers. Next, we design a comprehensive set of metrics to evaluate the reliability of these generated answers. The combined collection of the original nine datasets, the generated answers by LLMs, and the evaluation metrics is referred to as RelQA. RelQA is used to train a discriminator RelD.

\subsection{DATA COLLECTION}
RelQA consists of nine sub-datasets: SQuAD~\citep{rajpurkar2016squad}, DuReader~\citep{he2017dureader}, HotpotQA~\citep{yang2018hotpotqa}, MSMARCO~\citep{nguyen2016ms}, NewsQA~\citep{trischler2016newsqa}, QuAC~\citep{choi2018quac}, CoQA~\citep{reddy2019coqa}, TriviaQA-Web~\citep{joshi2017triviaqa}, and TriviaQA-Wikipedia~\citep{joshi2017triviaqa}. 
The detailed collecting steps are as follows:

\textbf{Step 1 (Dataset Selection):} 
These datasets are selected due to their unique characteristics, diverse sources, and the enrichment they bring to the overall collection. They cover extractive reading comprehension (ERC), multiple-choice (MC), and multi-turn dialogues (MTD) categories. They originate from sources such as Wikipedia, Baidu Zhidao, Bing search, and other platforms, while encompassing domains such as student education, news, web articles, and general knowledge. 

\textbf{Step 2 (Formatting and Integration):} 
To ensure compatibility and remove dataset boundaries, we perform formatting and integration for all selected datasets based on the aforementioned categories. 
Each dataset follows a specific standardized format, as illustrated in Table~\ref{tab:format} (the second column). 
We represent the datasets of all categories as $\{L_i,D_i\}$, where $L_i$ denotes a specific dataset and $D_i$ denotes its standardized format.

\textbf{Step 3 (Preprocessing):}
To facilitate effective processing and generation of answers, we employ preprocessing techniques on the dataset. This involves two primary aspects: personalized prompt instruction design and addressing the limitations associated with long texts. For personalized prompt instruction design, we create question-adaptive prompt instructions for each question based on the question type, as shown in Table~\ref{tab:format} (the third column). These prompt instructions guide LLMs in generating better answers that align with different types of questions. 
To address the challenge of long texts, we implement a sliding window approach~\citep{koay2021sliding}, segmenting the texts into smaller windows, each containing 4,000 tokens. This ensures that LLMs receive clear prompt instructions and can effectively handle texts of varying lengths, resulting in more accurate and contextually appropriate answers.

\begin{table*}[t]
\footnotesize
\caption{\footnotesize The format and prompt instuctions of three types of datasets. $a_i$: the answer in ERC or MTD, or the correct answer in MC. $a'_i$: the wrong answers in MC.}
    \begin{center}
        \begin{threeparttable}
        \resizebox{0.85\textwidth}{!}{
            \begin{tabular}{l|ll}
                \toprule
                    \bf Type&\bf Format &\bf Prompt instruction\\
                \midrule
                ERC&$D_i=\{c_i,q_i,a_i\}$&Given the following context $c_i$ and the question $q_i$. Please provide the answer.\\
                MC&$D_i=\{c_i,q_i,a_i,a'_i\}$&Given the following context $c_i$ and the question $q_i$. Please select the best answer from the candidate answers $\{a_i,a'_i\}$.\\
                MTD&$D_i=\{h_i,q_i,a_i\}$&Given the history conversation $h_i$ and the current question $q_i$. Please provide the answer.\\
                \bottomrule
            \end{tabular}}
        \end{threeparttable}
    \end{center}
    \label{tab:format}
\end{table*}

\textbf{Step 4 (Answer Generation):}
We employ several powerful LLMs, including LLaMA~\citep{touvron2023llama}, BLOOM~\citep{scao2022bloom}, GPT-J~\citep{gpt-j}, GPT-3~\citep{brown2020language}, and GPT-3.5~\footnote{\label{GPT-3.5}https://chat.openai.com/}, to generate answers for evaluation. In the case of longer texts, we slide the window over the text and generate outputs for each window. The generated outputs for each window are stored to facilitate subsequent filtering and selection of the optimal answers. To maintain answer stability, we ask an LLM to generate the answer three times for each question and select the majority answer as the final answer. Furthermore, to ensure the overall quality and reliability of the generated answers, we conduct quality assurance procedures, including automated checks to identify and re-generate incomplete sentences by detecting missing sentence-ending punctuation, among others.

\subsection{METRIC SELECTION}
To evaluate the reliability of LLMs' generated answers, it is crucial to select appropriate metrics that capture different aspects of answer quality. We employ four types of metrics, including LLM-assessment metric, human metric, machine metric, and composite metric, to comprehensively evaluate the generated answers.

\textbf{LLM-assessment metric} is inspired by the concept of LLMs' self-evaluation, where LLMs occasionally demonstrate the ability to assess their own output correctly without human intervention~\citep{chiang2023can,yan2023refining}. This metric comprises two specific indicators: the goodness of a generated answer and the similarity between the generated answer and the ground-truth answer. 
By obtaining the goodness score and similarity score of a generated answer, we can evaluate its quality and how closely it aligns with the ground-truth answer. Higher scores indicate better quality and semantic alignment. The LLM-assessment metric provides valuable insights into the LLMs' ability to evaluate the quality of generated answers.

\textbf{Human metric} plays a significant role in evaluating the LLM's performance from a human perspective. It includes a human score, which is a binary label assigned to each answer based on the degree of match between the LLM's generated answer and the ground-truth answer, along with the assigned goodness score. The human metric labeling is as follows:
i) When the LLM's generated answer is the same as the ground-truth answer and receives a goodness score of 4 or 5, the human metric is labeled as 1. This indicates that the LLM has successfully generated a correct and high-quality answer that aligns with the expected answer.
ii) When the LLM's generated answer is different from the ground-truth answer and receives a goodness score of 1, 2, or 3, the human metric is labeled as 2. 
This suggests that the LLM's generated answer is incorrect or of lower quality compared to the ground-truth answer.
iii) For cases where the LLM's generated answer neither matches the ground-truth answer nor falls within the aforementioned goodness score ranges, the human metric is labeled as 0. This label represents a neutral or ambiguous classification, indicating that the answer may require further examination or subjective judgment.
The human metric captures the human perception of the LLM's performance.

\textbf{Machine metric} draws inspiration from question-answering and dialogue systems, which rely on objective metrics to assess the quality of generated answers. It encompasses various categories, including accuracy metrics, overlap metrics, similarity metrics, and diversity metrics. Examples of machine metrics include F1 score, Recall, BLEU~\citep{papineni2002bleu}, BERT score~\citep{zhang2019bertscore}, ROUGE (ROUGE-1, ROUGE-2, ROUGE-L)~\citep{lin2004rouge}, Distinct-N (Distinct-1, Distinct-2)~\citep{li2015diversity}, Greedy matching, and Embedding scores (average, extreme)~\citep{liu2016not}. 
Specifically, accuracy metrics assess the correctness of generated answers compared to the ground truth, including F1 score.
Overlap metrics measure the overlap between generated answers and the ground truth, including BLEU, Recall, ROUGE.
Similarity metrics capture the semantic similarity between generated answers and the ground truth, including BERT score, Greedy matching and Embedding scores (average, extreme).
Diversity metrics measure the diversity of the generated answers, including Distinct-N.
These metrics objectively evaluate the semantic alignment, relevance, diversity, and quality of generated answers, enabling a comprehensive assessment of LLMs' answers.

\textbf{Composite metric} is designed to provide a comprehensive evaluation of a model's performance by combining multiple aspects. It includes a final score and a final tag to summarize the evaluation. Each of the metrics mentioned above contributes to the final score, with specific emphasis given to certain metrics. For instance, Recall and ROUGE (ROUGE-1, ROUGE-2, ROUGE-L) may be assigned higher weights (e.g., twice the weight) to highlight the importance of maintaining information~\citep{lyu2022attention,lyu2023backdoor}. The weights of different metrics can be dynamically optimized to better assess their importance in real-world scenarios as demonstrated in Experiment~\ref{Optimizing}. 
The final tag is a binary label assigned based on the average score. If the average score is greater than 0.5, it is labeled as 1; otherwise, it is labeled as 0. The final tag simplifies the evaluation outcome, indicating whether the LLMs' generated answer is considered reliable or not. In summary, these metrics collectively evaluate the quality of answers generated by LLMs compared to the ground-truth answers.

\subsection{DATA EXPLORATORY ANALYSIS}
\begin{figure}[!t]
  \centering
  \includegraphics[width=0.72\linewidth]{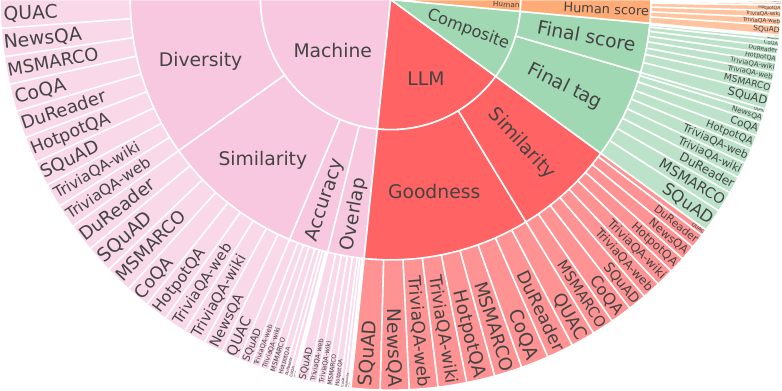}
  \caption{\footnotesize A data exploratory analysis of the constructed RelQA based on different metrics.}
  \label{fig:data}
  \vspace{-1em}
\end{figure}

In this section, we conduct a data exploratory analysis of the constructed RelQA dataset, which comprises a total of 1,372,130 samples, including generated answers by five selected LLMs. Among these, 743,910 samples are assigned as reliable and 628,220 samples as unreliable based on the final tag metric. We divide the possible ranges of all metrics into three equal parts, representing low, medium, and high levels. Fig~\ref{fig:data} illustrates the distribution of each dataset at the high level for each metric.
%
%
We also present the distributions of different datasets among various metrics as shown in 
Table~\ref{tab:llm-ass}, Table~\ref{tab:human}, Table~\ref{tab:mach} and 
Table~\ref{tab:comp}.

\begin{table}[!t]
\footnotesize
\caption{\footnotesize The distribution of each dataset in RelQA on LLM-assessment metric.}
    \begin{center}
        \begin{threeparttable}
        \resizebox{0.4\textwidth}{!}{
            \begin{tabular}{l|ccc|ccc}
                \toprule
                \multirow{2}{2cm}{\bf Dataset} & 
               \multicolumn{3}{c|}{\bf Goodness} &
                \multicolumn{3}{c}{\bf Similarity}\\
                        &\bf Low&\bf Medium&\bf High
                        &\bf Low&\bf Medium&\bf High\\
                \midrule
                SQuAD	&0.11\%		&0.42\%		&99.47\%	&	33.71\%		&2.50\%		&63.8\%	\\
DuReader	&2.77\%		&5.60\%		&91.63\%	&	15.73\%		&34.01\%	&	50.26\%	\\
HotpotQA	&1.47\%		&1.35\%		&97.18\%		&37.57\%		&5.52\%		&56.9\%\\
MSMARCO	&1.62\%	&2.43\%	&95.95\%	&13.58\%&	11.53\%	&74.89\\
NewsQA	&0.66\%	&0.91\%	&98.43\%	&21.67\%	&25.44\%	&52.89\\
QUAC	&8.87\%	&8.41\%	&82.72\%	&60.28\%	&18.3\%	&21.41\\
CoQA	&1.37\%	&3.08\%	&95.55\%	&18.45\%	&7.43\%	&74.13\\
TriviaQA-web	&1.25\%	&0.63\%	&98.12\%	&31.18\%	&6.16\%	&62.66\\
TriviaQA-wiki	&1.36\%	&0.66\%	&97.99\%	&31.36\%	&6.54\%	&62.11\\
                \bottomrule
            \end{tabular}}
        \end{threeparttable}
    \end{center}
    \label{tab:llm-ass}
    \vspace{-1em}
\end{table}

\begin{table}[!t]
\footnotesize
\caption{\footnotesize The distribution of each dataset in RelQA on Human metric.}
    \begin{center}
        \begin{threeparttable}
        \resizebox{0.32\textwidth}{!}{
            \begin{tabular}{l|ccc}
                \toprule
                \multirow{3}{2cm}{\bf Dataset} & 
                \multicolumn{3}{c}{\bf Human score}\\
                        &\bf Reliable&\bf Unreliable&\bf Ambiguous\\
                \midrule
        SQuAD	&32.79\%	&0.49\%	&	66.71\%	\\
        DuReader		&0.42\%	&8.31\%		&91.27\%	\\
        HotpotQA	&19.75\%	&	2.73\%		&77.52\%	\\
        MSMARCO	&6.95\%		&3.99\%		&89.06\%	\\
        NewsQA	&2.09\%		&1.53\%		&96.38\%	\\
        QUAC	&0.81\%		&17.16\%		&82.03\%	\\
        CoQA	&8.08\%		&4.22\%		&87.71\%	\\
        TriviaQA-web	&25.77	&1.75&	72.49\%	\\
        TriviaQA-wiki	&24.29	&1.87&	73.84\%	\\
                \bottomrule
            \end{tabular}}
        \end{threeparttable}
    \end{center}
    \label{tab:human}
    \vspace{-1em}
\end{table}

\begin{table*}[!t]
\footnotesize
\caption{\footnotesize The distribution of each dataset in RelQA on Machine metric.}
    \begin{center}
        \begin{threeparttable}
        \resizebox{0.8\textwidth}{!}{
            \begin{tabular}{l|ccc|ccc|ccc|ccc}
                \toprule
                \multirow{2}{2cm}{\bf Dataset} & 
               \multicolumn{3}{c|}{\bf Accuracy} &
               \multicolumn{3}{c|}{\bf Overlap} &
               \multicolumn{3}{c|}{\bf Similarity} &
                \multicolumn{3}{c}{\bf Diversity}\\
                        &\bf Low&\bf Medium&\bf High
                        &\bf Low&\bf Medium&\bf High
                        &\bf Low&\bf Medium&\bf High
                        &\bf Low&\bf Medium&\bf High\\
                \midrule
                SQuAD&	25.27\%&	30.03\%&	44.69\%&	32.47\%&	25.95\%&	41.58\%&	0.19\%&	5.10\%&	94.71\%&	0.00\%&	13.45\%&	86.55\%\\
                DuReader&	49.35\%	&34.53\%	&16.12\%	&56.51\%	&30.81\%	&12.67\%	&0.13\%	&3.98\%	&95.89\%	&0.03\%	&5.86\%	&94.10\%\\
                HotpotQA	&53.79\%	&21.49\%	&24.73\%	&60.26\%	&15.91\%	&23.83\%	&0.38\%	&20.57\%	&79.06\%	&0.00\%	&9.61\%	&90.39\%\\
                MSMARCO	&33.99\%	&35.91\%	&30.09\%	&37.69\%	&35.92\%	&26.38\%	&0.19\%&6.40\%	&93.41\%		&0.00\%	&5.03\%	&94.97\%\\
                NewsQA	&70.53\%	&22.92\%	&6.56\%	&75.28\%	&19.04\%	&5.68\%	&1.52\%	&31.88\%	&66.60\%	&0.00\%	&3.17\%	&96.83\%\\
                QUAC	&85.63\%	&9.83\%	&4.54\%	&87.52\%	&8.59\%	&3.89\%	&0.51\%	&35.36\%	&64.13\%&	0.01\%	&2.22\%	&97.77\%\\
                CoQA	&56.09\%	&28.10\%	&15.81\%	&64.49\%	&22.06\%	&13.46\%	&0.54\%	&18.22\%	&81.24\%&	0.00\%	&5.77\%	&94.23\%\\
                TriviaQA-web	&48.26\%	&20.93\%	&30.8\%	&54.17\%	&15.49\%	&30.34\%&	1.00\%	&20.74\%	&78.26\%	&0.00\%	&17.88\%&82.12\%\\
                TriviaQA-wiki	&47.83\%	&21.71\%	&30.46\%	&53.56\%	&16.40\%	&30.05\%	&1.06\%	&21.75\%	&77.19\%	&0.01\%	&17.73\%	&82.26\%\\
                \bottomrule
            \end{tabular}}
        \end{threeparttable}
    \end{center}
    \label{tab:mach}
\end{table*}

\begin{table}[!t]
\footnotesize
\caption{\footnotesize The distribution of each dataset in RelQA on Composite metric.}
    \begin{center}
        \begin{threeparttable}
        \resizebox{0.42\textwidth}{!}{
            \begin{tabular}{l|ccc|cc}
                \toprule
                \multirow{2}{2cm}{\bf Dataset} & 
               \multicolumn{3}{c|}{\bf Final score} &
                \multicolumn{2}{c}{\bf Final tag}\\
                        &\bf Low&\bf Medium&\bf High
                        &\bf Reliable&\bf Unreliable\\
                \midrule
SQuAD&	3.56\%&	44.01\%&	52.43\%&	78.57\%&	21.43\%\\
DuReader&	6.32\%&	67.85\%&	25.83\%&	58.57\%&	41.43\%\\
HotpotQA&	15.89\%&	56.02\%&	28.10\%&	47.75\%&	52.25\%\\
MSMARCO&	5.52\%&	51.88\%&	42.60\%&	72.29\%&	27.71\%\\
NewsQA&	27.65\%&	60.55\%&	11.80\%&	33.15\%&	66.85\%\\
QUAC&	50.08\%&	43.04\%&	6.88\%&	16.44\%&	83.56\%\\
CoQA&	15.43\%&	62.79\%&	21.78\%&	45.75\%&	54.25\%\\
TriviaQA-web&	17.83\%&	49.50\%&	32.67\%&	53.34\%&	46.66\%\\
TriviaQA-wiki&	19.32\%&	48.36\%&	32.33\%&	53.41\%&	46.59\%\\
                \bottomrule
            \end{tabular}}
        \end{threeparttable}
    \end{center}
    \label{tab:comp}
    \vspace{-1em}
\end{table}

First, we analyze the differences in the LLM-assessment metric across different datasets. Regarding the ``goodness'' metric, the QUAC dataset performs poorly in terms of answer quality, with a high score percentage of 82.72\%, while the SQuAD dataset excels in generating high-quality answers, with a high score percentage of 99.47\%. Other datasets generally achieve high score percentages above 90\%.
Regarding the ``similarity'' metric, the MSMARCO dataset demonstrates the highest similarity to the reference answers, with a high similarity percentage of 74.89\%. Conversely, the QUAC dataset also performs poorly in terms of similarity, with a low similarity percentage of 60.28\%. 

Next, we analyze the differences in the human metric across different datasets. The proportions of reliable evaluations vary significantly in the ``human score'' metric. The lowest proportion is 0.42\% for DuReader-master, while the highest is 32.79\% for SQuAD. Similarly, the proportions of unreliable evaluations differ, with the lowest being 0.49\% for SQuAD and the highest being 17.16\% for QUAC. Additionally, the proportion of ambiguous evaluations is highest for newsQA at 96.38\% and lowest for QUAC at 66.71\%. 

Afterwards, we analyze the differences in the machine metric across different datasets. In terms of ``accuracy metrics'', the QUAC dataset performs the worst, with a high score percentage of only 4.54\%. The high score percentages for other datasets range between 4.54\% and 30.8\%, with a median around 20\%. 
In terms of ``overlap metrics'', the QUAC dataset also performs poorly in terms of low overlap, with a low score percentage of 87.52\%. The low score percentages for other datasets range from 32.47\% to 75.28\%, with no significant high scores observed overall.
Regarding ``similarity metrics'', DuReader, SQuAD, and MSMARCO perform well in terms of high similarity scores, with the highest scores being 95.89\%, 94.71\%, and 93.41\% respectively. In contrast, newsQA and QUAC exhibit lower similarity scores, with the highest scores being 66.6\% and 64.13\% respectively. 
Notably, there are consistencies between the similarity scores in machine metrics and the similarity scores in LLM-assessment metrics.
In ``diversity metrics'', QUAC, newsQA, and MSMARCO perform well in terms of high diversity scores, with the highest scores being 97.77\%, 96.83\%, and 94.97\% respectively. This is likely due to the higher question diversity in these datasets, allowing models to exhibit more creativity and diversity in generating answers. Other datasets also maintain high diversity scores, all above 80\%.

Finally, we analyze the differences in composite evaluation metrics across different datasets. In terms of the ``final score'' metric, the QUAC dataset performs the worst, with a high composite score percentage of 6.88\%. Conversely, the SQuAD dataset achieves the highest composite score, with a high percentage of 52.43\%. It is evident that none of the datasets achieve particularly high composite scores.
In terms of the ``final tag'' metric, the SQuAD dataset exhibits the highest proportion indicating answer reliablity, at 78.57\%, while the QUAC dataset has the lowest proportion at 16.44\%. This aligns with the human metric, as the SQuAD dataset primarily consists of simple extractive reading comprehension, making it easier for models to generate reliable answers. On the other hand, QUAC involves open-domain dialogue with more complex semantic understanding, posing challenges for models to generate reliable answers.

\section{DISCRIMINATOR}
\begin{figure}[!t]
  \centering
  \includegraphics[width=0.92\linewidth]{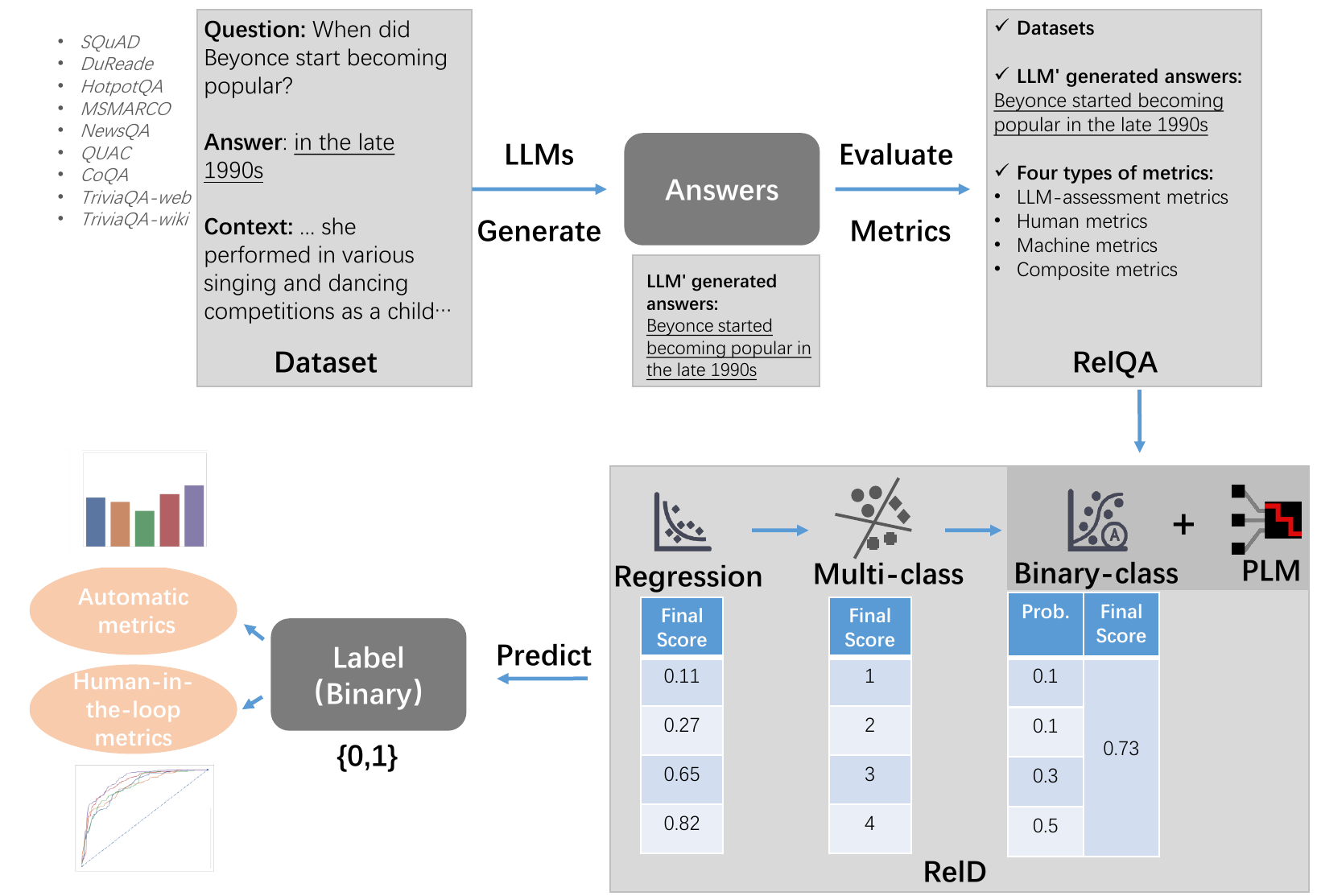}
  \caption{\footnotesize The process of building the discriminator RelD, which is trained on the constructed dataset RelQA and used to detect hallucination of LLMs' generated answers.}
  \label{fig:framework}
  \vspace{-1em}
\end{figure}

In this section, we introduce a novel and robust discriminator called RelD, which is designed to assess the reliability of answers generated by LLMs. To ensure that RelD closely aligns with human evaluation, we employ an appropriate method to train RelD and make it fit the final score based on human evaluation. The process of constructing RelD is illustrated in Fig.~\ref{fig:framework}.

\subsection{REGRESSION TO MULTI-CLASS CLASSIFICATION}
Initially, we employ a regression approach to train the discriminator RelD in order to fit the final score and align with human evaluation. However, our experiments reveal that the regression approach performs poorly, possibly due to the use of the mean square error loss function. Consequently, we convert the regression task into a classification task to improve the fitting.
Specifically, 
In this process, we normalize the final score into different numbers of classes, such as four, six, eight, and ten, for multi-class classification. For instance, we assign the first category in a four-category classification to final scores ranging from 0 to 0.25. After experiments as shown in Sec.~\ref{categories}, we ultimately choose a ten-class classification approach.
The theoretical foundation of this method mainly lies in information theory and the cross-entropy loss function. Cross-entropy is a common information theory measure used to quantify the distance between two probability distributions. In the case of multi-classification problems, the cross-entropy loss function is defined as follows:
\begin{gather}
\footnotesize
L = - \sum (y_i \cdot \log(p_i)),
\vspace{-1em}
\end{gather}
where $y_i$ represents the true label of the $i$-th category, and $p_i$ represents the predicted probability of the $i$-th category by the discriminator RelD. Our objective is to minimize this loss function during the training of RelD.
In practice, we employ the softmax function to convert the original output of RelD into a probability distribution. 
 
One potential advantage of this method is that the classification task, which focuses on distinguishing different categories, may facilitate capturing subtle differences among the final scores. Furthermore, the cross-entropy loss function exhibits greater stability compared to the mean square error loss function when dealing with imbalanced datasets. However, it is important to note that in certain situations, multi-class tasks may introduce overly complex information, leading to a notable disparity between the concepts learned by the discriminator and human intuitive perception. For example, dividing a problem into five categories, such as ``not reliable'', ``weakly reliable'', ``moderately reliable'', ``strongly reliable'' and ``highly reliable'', may surpass most people's intuitive understanding of the fundamental categories of ``reliable'' and ``unreliable''.

\subsection{MULTI-CLASS TO BINARY-CLASS CLASSIFICATION}
Based on the aforementioned analysis, we further convert the multi-class task into a binary classification task, which may better align with human intuitive perception.
Here, we present three possible approaches for this conversion, each with its theoretical support and definition:

\textbf{Normalization.} This method is based on threshold decision theory. It involves converting all class information into binary labels by directly normalizing the final score to 0 and 1, which serves as the final probability value for classification. However, this approach may result in some information loss as continuous scores are transformed into discrete classes.

\textbf{Discrete Values.} This method is grounded in maximum likelihood estimation, a commonly used parameter estimation technique in statistics. Here, we consider the highest predicted probability from the discriminator as the final probability value for classification. For example, in a four-class classification scenario, if the probabilities corresponding to the classes are 0.1, 0.1, 0.1, and 0.7, respectively, we would use 0.7 as the final probability value. The advantage of this method lies in its simplicity, although the drawback is that we do not know which class the maximum probability value corresponds to.

\textbf{Weighted Average Probability.} The theoretical basis for this method stems from decision theory, particularly the concept of expected utility, which involves taking a weighted average of all possible outcomes and their corresponding utilities (in this case, predicted probabilities). The goal of this approach is to determine a weighted average value that best represents the predicted probabilities for each class from the discriminator. In this method, we multiply the probability of each class predicted by the discriminator with its corresponding weight, summing them up to obtain a final probability value. This value can then be used for binary classification tasks. The formula for this method is as follows:
\begin{gather}
\footnotesize
p_i' = \frac{(\sum w_i \cdot p_i) - w_{\text{min}}}{w_{\text{max}} - w_{\text{min}}},
\vspace{-1em}
\end{gather}
where $p_i$ represents the probability output of the discriminator for class $i$, $w_i$ denotes the weight for class $i$, and $w_{\text{min}}$ and $w_{\text{max}}$ are the minimum and maximum weights, respectively. We set the threshold to 0.5 and use the cross-entropy loss function for approximation. It allows for a more refined fitting of regression tasks and has demonstrated better performance compared to the previous two methods, as indicated by Sec.~\ref{probability}.

\subsection{Backbone of the Discriminator}
We utilize a Pre-trained Language Model (PLM), such as ELECTRA~\citep{ELECTRA}, as the backbone of the discriminator RelD. Through our experiments, we have demonstrated that ELECTRA outperforms other PLMs, including BERT~\citep{BERT}, RoBERTa~\citep{RoBERTa}, and DeBERTa~\citep{DeBERTa}, as indicated in Section~\ref{Backbone}. RelD takes questions along with contexts and LLMs' generated answers as input, generating a classification label to determine the reliability of a generated answer. It uses the weighted average probability approach to fit the ground-truth answers.

\section{EXPERIMENTS}
In this section, we conduct experiments to evaluate the effectiveness of RelD in detecting the reliability of LLMs' generated answers using both automatic metrics and human-in-the-loop metrics. 

\subsection{EXPERIMENTAL SETUP}
The experiments are conducted using TESLA A100 GPUs for answer generation and GTX 3090 GPUs for training RelD with PyTorch in Python. During the training of RelD, we set the batch size to 32 and the sequence length to 128. Hyperparameters such as weight decay (0.01), $\beta_1$ (0.9), and $\beta_2$ (0.999) are maintained. The learning rate is set to 2e-05. We train RelD for 20 epochs.

\textbf{Baselines and metrics. }
We validate the effectiveness of the proposed RelD on well-known LLMs, including LLaMA (LLaMA-7B)\citep{touvron2023llama}, BLOOM (BLOOM-7B)\citep{scao2022bloom}, GPT-J (GPT-J-6B)\citep{gpt-j}, GPT-3\citep{brown2020language}, and GPT-3.5~$^{\ref{GPT-3.5}}$. To evaluate the performance of RelD, we use accuracy (ACC) as the automatic metrics and ROC curve analysis with the area under the ROC curve (AUC) as the human-in-the-loop metrics.
The automatic evaluation process utilizes the final tag as the ground-truth label, while the human-in-the-loop evaluation involves human ratings as the ground-truth labels.
Specifically, we randomly select 9,000 QA pairs, with 1,000 from each dataset in RelQA, for human ratings. We enroll nine volunteers and divide them into three groups to ensure evaluation stability. Each group provides scores of 0 or 1 for the randomly selected 3,000 QA pairs. Inter-rater agreement is calculated using Krippendorff's Alpha (IRA) to ensure the confidence of the human ratings. For controversial ratings with low agreement ($<$0.7), we discard the corresponding QA pair and replace it with another.



\subsection{MAIN RESULTS}
We conduct experiments to evaluate the effectiveness of the proposed RelD as follows:

\textbf{Experiment 1: RelD's Performance across Different LLMs.} We conduct ten-fold cross-validation and report the average performance on the validation dataset. Based on the results presented in Table~\ref{tab:exp1}, it's observed that both the automatic and human-in-the-loop evaluations consistently exceed 0.8 for all LLMs, with minimal variation between different models (p<0.01). The strong correlation between the automatic and human-in-the-loop evaluations (p<0.01) suggests that the automatic scoring of the RelQA dataset could largely replace human scoring. It also indicates the robustness of RelD in detecting the reliability of different LLMs.

\begin{table}[!t]
\caption{\footnotesize Performance of RelD among the selected LLMs on the validation dataset.}
    \begin{center}
        \begin{threeparttable}    
        \resizebox{0.37\textwidth}{!}{
            \begin{tabular}{c|lllll}
                \toprule
                \bf LLM &LLaMA&BLOOM&GPT-J&GPT-3&GPT-3.5\\
                \midrule
                Automatic &0.855&0.846&0.827&0.863&0.881\\
                Human&0.826&0.830&0.835&0.869&0.894\\
                \midrule
                Average score &0.841&	0.838	&0.831&	0.866&	0.888\\
                \bottomrule
                
            \end{tabular}}
        \end{threeparttable}
    \end{center}
    \label{tab:exp1}
    \vspace{-1em}
\end{table}

\begin{figure}[!t]
  \centering
  \includegraphics[width=0.82\linewidth]{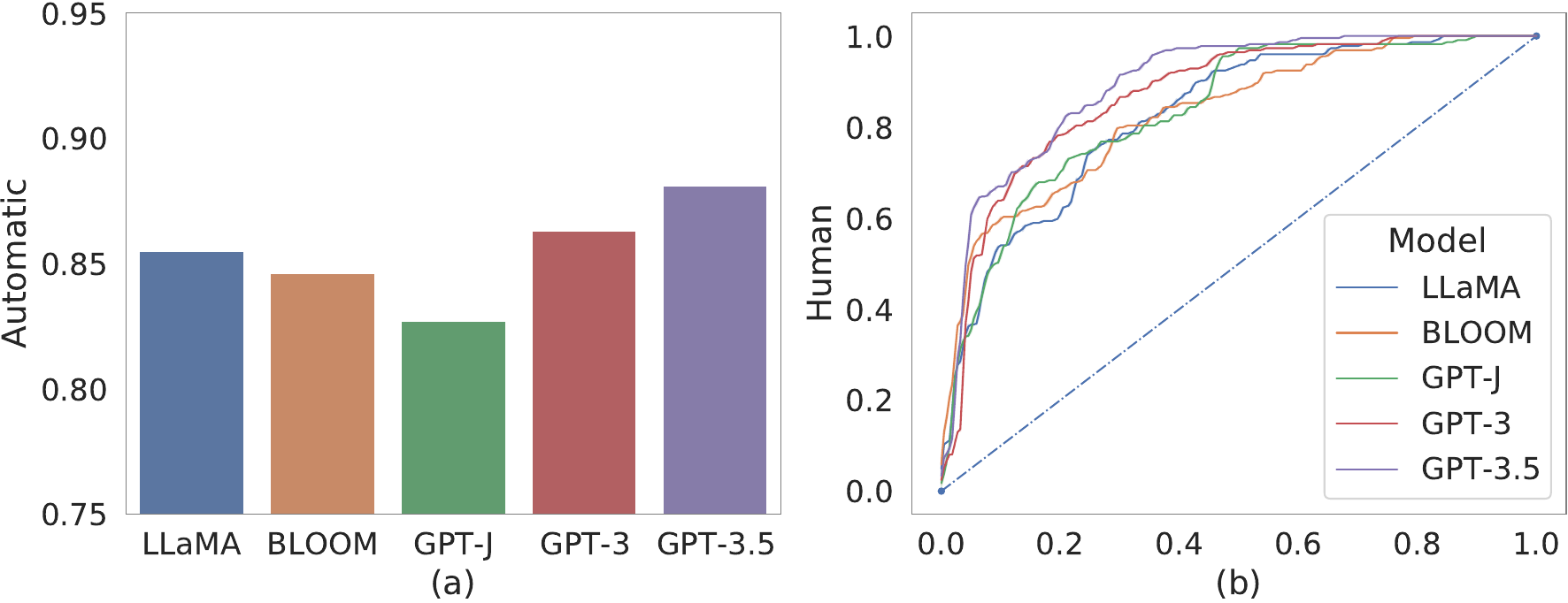}
  \caption{\footnotesize The visualization of RelD's performance among the selected LLMs on the validation dataset, including both automatic and human-in-the-loop metrics.}
  \label{fig:exp1}
  \vspace{-1em}
\end{figure}

\begin{figure}[!t]
  \centering
  \includegraphics[width=\linewidth]{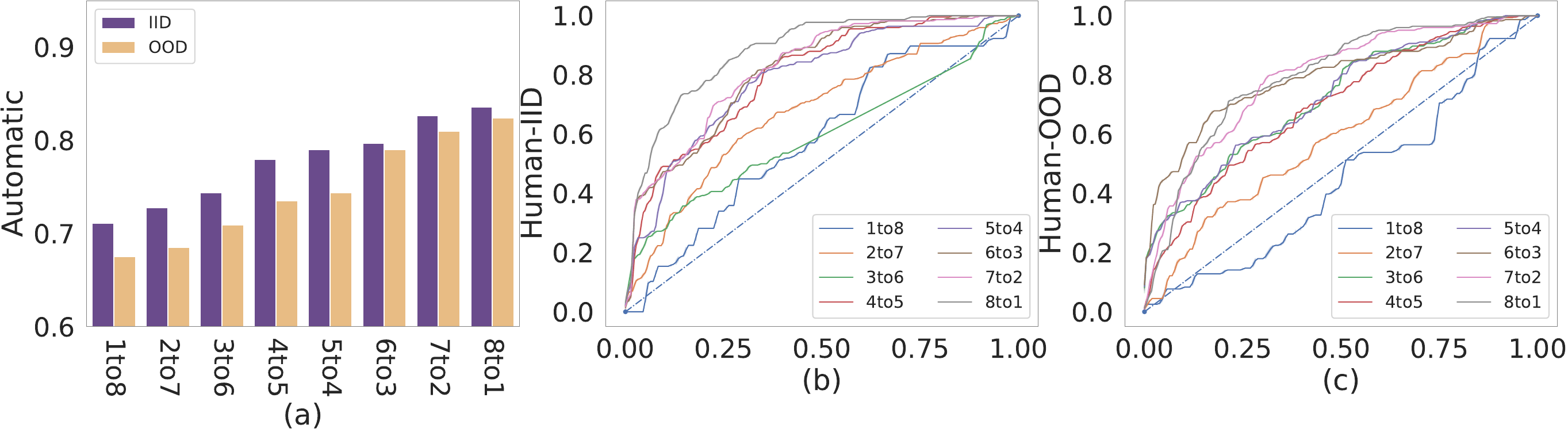}
  \caption{\footnotesize Performance of RelD on automatic metrics (a) and human-in-the-loop metrics (b)(c), including results on IID validation dataset (b) and OOD dataset (c) among the selected LLMs.}
  \label{fig:exp2}
  \vspace{-1em}
\end{figure}

\textbf{Experiment 2: RelD's Performance on IID and OOD Datasets}
We evaluate the performance of RelD on both In-distribution (IID) and Out-of-distribution (OOD) datasets. We randomly assign nine datasets from RelQA to the IID and OOD sets in various ratios, such as 1:8, 2:7, 3:6, and 4:5, and vice versa. For example, we train on 8 datasets and validate on 1 dataset. To ensure a balanced quantity of data in both the IID and OOD sets, we perform downsampling by randomly selecting 3,000 samples from each dataset.
Considering that different datasets acting as IID or OOD may yield different results, we conduct five experiments for each ratio group and provide average values along with the range of error. This approach allows us to accurately assess the generalization ability of RelD.
To evaluate the performance on the IID dataset, we use 30\% of the IID data as a validation dataset. For the OOD evaluation, we directly test RelD on the entire OOD dataset. The results are presented in Table~\ref{tab:exp2} and Fig.~\ref{fig:exp2}. We observe that when the IID ratio is set to 5 or higher, RelD consistently achieves automatic and human-in-the-loop evaluations above 0.7 on both the IID and OOD datasets. This indicates that RelD exhibits a strong generalization capability in handling OOD data as well as alignment with human evaluation predictions.



\begin{table*}[t]
\footnotesize
\caption{\footnotesize Performance of RelD on IID and OOD datasets. IID results are based on a 30\% validation dataset from the IID dataset, while OOD results are obtained from the entire OOD dataset.}
    \begin{center}
        \begin{threeparttable}
        \resizebox{0.8\textwidth}{!}{
            \begin{tabular}{c|c|c|ccccccccc}
                \toprule
                \bf LLM
                &\bf Metrics
                &\bf Distribution
                &\bf 1 to 8 &\bf 2 to 7 &\bf 3 to 6 &\bf 4 to 5 &\bf 5 to 4 &\bf 6 to 3&\bf 7 to 2&\bf 8 to 1&\bf Average\\	
                \midrule
                \bf \multirow{4}{*}{\bf LLaMA}&
                \multirow{2}{*}{\bf Automatic}
                &IID
&0.698$_{±.021}$ &0.723$_{±.018}$	&0.762$_{±.029}$ &0.785$_{±.016}$	&0.776$_{±.012}$	&0.806$_{±.022}$	&0.821$_{±.014}$	&0.832$_{±.010}$ &0.775$_{±.018}$
\\
                &&OOD
&0.672$_{±.023}$	&0.675$_{±.020}$ &0.701$_{±.017}$&	0.747$_{±.011}$	&0.735$_{±.028}$	&0.798$_{±.026}$&0.815$_{±.024	}$	&0.820$_{±.013}$	&0.745$_{±.020}$	
\\
                &\multirow{2}{*}{\bf Human}
                &IID
&0.550$_{±.019}$&0.693$_{±.027}$	&0.721$_{±.015	}$	&0.758$_{±.010}$	&0.763$_{±.023	}$	&0.791$_{±.018	}$	&0.839$_{±.011	}$	&0.862$_{±.025}$	&0.747$_{±.019}$	
\\
                &&OOD
&0.487$_{±.021	}$	&0.547$_{±.017	}$	&0.585$_{±.029	}$	&0.634$_{±.022	}$	&0.732$_{±.015	}$	&0.748$_{±.014	}$	&0.73$_{±.027	}$	&0.744$_{±.012}$	&0.651$_{±.020}$	
\\
                \midrule
                \bf \multirow{4}{*}{\bf BLOOM}&
                \multirow{2}{*}{\bf Automatic}
                &IID
&0.701$_{±.024	}$	&0.729$_{±.026}$	&0.755$_{±.013}$	&0.790$_{±.017}$	&	0.777$_{±.020	}$	&0.801$_{±.028	}$	&0.817$_{±.016	}$	&0.827$_{±.011}$	&0.775$_{±.019}$	
\\
                &&OOD
&0.674$_{±.018}$	&0.678$_{±.021	}$	&0.705$_{±.012	}$	&0.750$_{±.010	}$	&0.739$_{±.019	}$	&0.799$_{±.016	}$	&0.817$_{±.023	}$	&0.822$_{±.015}$	&0.748$_{±.017}$	
\\
                &\multirow{2}{*}{\bf Human}
                &IID
&0.539$_{±.013	}$	&0.680$_{±.014	}$	&0.695$_{±.028	}$	&0.747$_{±.019	}$	&0.759$_{±.022	}$	&0.778$_{±.024	}$	&0.834$_{±.012	}$	&0.854$_{±.011}$	&0.736$_{±.018}$	
\\
                &&OOD
&0.462$_{±.0120}$	&	0.521$_{±.011	}$	&0.546$_{±.025}$	&	0.628$_{±.016	}$	&0.731$_{±.023	}$	&0.725$_{±.017	}$	&0.732$_{±.020	}$	&0.725$_{±.027}$	&0.634$_{±.019}$	
\\
                \midrule
                \multirow{4}{*}{\bf GPT-J}&
                \multirow{2}{*}{\bf Automatic}
                &IID
&0.673$_{±.027	}$	&0.710$_{±.015	}$	&0.757$_{±.016	}$	&0.765$_{±.014}$	&	0.788$_{±.011	}$	&0.810$_{±.029	}$	&0.831$_{±.021	}$	&0.830$_{±.018}$	&0.771$_{±.019}$	
\\
                &&OOD
&0.685$_{±.016	}$	&0.677$_{±.012	}$	&0.706$_{±.022	}$	&0.746$_{±.020	}$	&0.733$_{±.011	}$	&0.795$_{±.017	}$	&0.812$_{±.014	}$	&0.810$_{±.026}$	&0.746$_{±.017}$	
\\
                &\multirow{2}{*}{\bf Human}
                &IID
&0.556$_{±.019	}$	&0.660$_{±.010}$	&	0.698$_{±.015	}$	&0.726$_{±.024	}$	&0.759$_{±.022	}$	&0.778$_{±.013	}$	&0.804$_{±.021	}$	&0.850$_{±.018}$	&0.729$_{±.018}$	
\\
                &&OOD
&0.451$_{±.026	}$	&0.523$_{±.013	}$	&0.557$_{±.024	}$	&0.605$_{±.012	}$	&0.731$_{±.011	}$	&0.725$_{±.020	}$	&0.733$_{±.028	}$	&0.721$_{±.023}$	&0.631$_{±.020}$	
\\
                \midrule
                \multirow{4}{*}{\bf GPT-3}&
                \multirow{2}{*}{\bf Automatic}
                &IID
&0.706$_{±.020	}$	&0.716$_{±.018	}$	&0.768$_{±.019	}$	&0.780$_{±.010	}$	&0.769$_{±.013	}$	&0.809$_{±.017	}$	&0.825$_{±.021	}$	&0.826$_{±.016}$	&0.775$_{±.017}$	
\\
                &&OOD
&0.681$_{±.015	}$	&0.680$_{±.014	}$	&0.710$_{±.010	}$	&0.753$_{±.011	}$	&0.729$_{±.026	}$	&0.792$_{±.019	}$	&0.813$_{±.012	}$	&0.815$_{±.024}$	&0.747$_{±.016}$	
\\
                &\multirow{2}{*}{\bf Human}
                &IID
&0.527$_{±.028	}$	&0.645$_{±.016}$	&	0.731$_{±.010}$	&0.745$_{±.023	}$	&0.782$_{±.017	}$	&0.793$_{±.026	}$	&0.836$_{±.015	}$	&0.897$_{±.013}$	&0.745$_{±.019}$	
\\
                &&OOD
&0.468$_{±.024	}$	&0.568$_{±.018	}$	&0.612$_{±.011	}$	&0.619$_{±.020	}$	&0.720$_{±.013	}$	&0.775$_{±.012	}$	&0.756$_{±.019	}$	&0.728$_{±.014}$	&0.656$_{±.016}$	
\\
                \midrule
                \multirow{4}{*}{\bf GPT-3.5}&
                \multirow{2}{*}{\bf Automatic}
                &IID
&0.711$_{±.010	}$	&0.728$_{±.012	}$	&0.744$_{±.015	}$	&0.780$_{±.014	}$	&0.790$_{±.018	}$	&0.797$_{±.027	}$	&0.827$_{±.010}$	&	0.836$_{±.021}$	&0.777$_{±.016}$	
\\
                &&OOD
&0.675$_{±.016	}$	&0.685$_{±.024}$	&	0.709$_{±.010	}$	&0.735$_{±.017	}$	&0.744$_{±.020	}$	&0.790$_{±.028	}$	&0.810$_{±.016	}$	&0.824$_{±.011}$	&0.747$_{±.018}$	
\\
                &\multirow{2}{*}{\bf Human}
                &IID
&0.586$_{±.027	}$	&0.677$_{±.012	}$	&0.746$_{±.013	}$	&0.797$_{±.014	}$	&0.786$_{±.018	}$	&0.812$_{±.023	}$	&0.821$_{±.012	}$	&0.880$_{±.011}$	&0.763$_{±.016}$	
\\
                &&OOD
&0.445$_{±.019	}$	&0.592$_{±.015	}$	&0.721$_{±.017	}$	&0.722$_{±.026	}$	&0.723$_{±.024	}$	&0.791$_{±.010}$	&	0.791$_{±.016	}$	&0.795$_{±.012}$	&0.698$_{±.017}$	\\
                \bottomrule
            \end{tabular}}
        \end{threeparttable}
    \end{center}
    \label{tab:exp2}
\end{table*}

\subsection{ABLATION STUDY}
After that, we conduct several experiments to evaluate the effectiveness of different modules in the proposed RelD. All results are performed on the validation dataset using ten-fold cross-validation.

\begin{figure}[!t]
  \centering
  \includegraphics[width=0.75\linewidth]{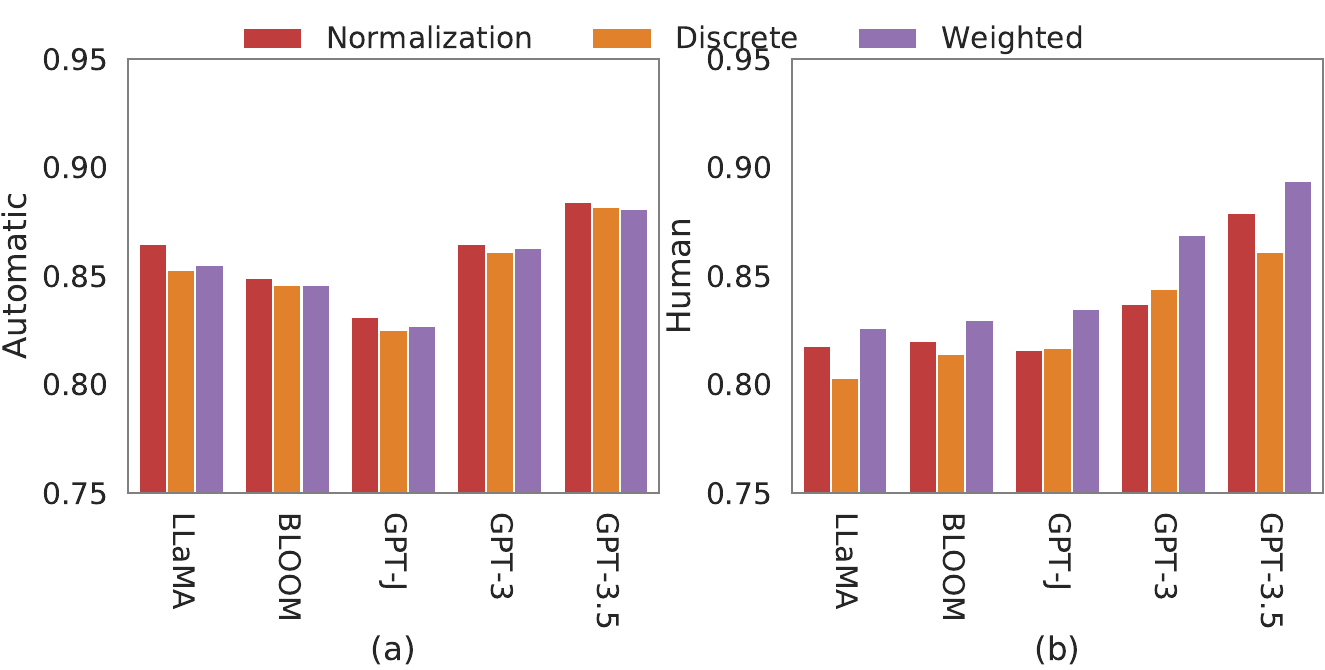}
  \caption{\footnotesize The performance of using weighted average probability is compared with using normalization and discrete values in automatic metrics (a) and human-in-the-loop metrics (b) on the validation dataset among the selected LLMs.}
  \label{fig:exp3}
  \vspace{-1em}
\end{figure}

\textbf{Experiment 3: Effectiveness of Weighted Average Probability. }
\label{probability}
We compare the performance of using normalization, discrete values, and weighted average probability in the conversion from multi-class to binary-class classification in both automatic and human-in-the-loop metrics. The results are presented in Fig.~\ref{fig:exp3}.
We observe that while using weighted average probability slightly underperforms normalization in terms of automatic metrics, it significantly outperforms normalization and discrete values in human-in-the-loop metrics across all LLMs. Therefore, we adopt weighted average probability as it offers a more intuitive and aligned approach from a human perspective.

\begin{figure}[!t]
  \centering
  \includegraphics[width=0.82\linewidth]{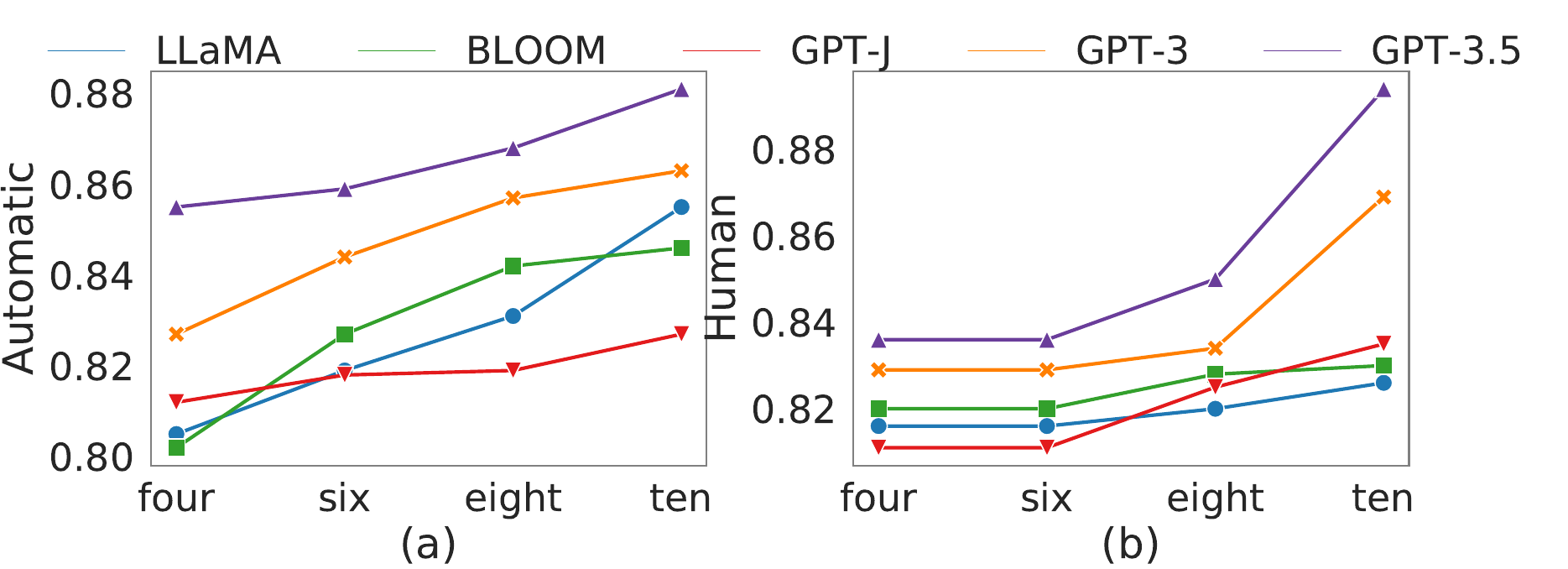}
  \caption{\footnotesize The performance of different numbers of categories in automatic (a) and human-in-the-loop metrics (b) on the validation dataset among the selected LLMs.}
  \label{fig:exp4}
  \vspace{-1em}
\end{figure}

\textbf{Experiment 4: Optimal Number of Categories. }
\label{categories}
We investigate the impact of the number of categories when converting regression into multi-class classification. We test four categories, six categories, eight categories, and ten categories. The results are shown in Fig.~\ref{fig:exp4}. It is evident that a higher number of categories leads to improved performance in human-in-the-loop metrics. This suggests that a larger number of categories brings the classification task closer to regression and enhances alignment with human cognition. Consequently, we ultimately convert the regression task into a ten-category classification task and then discern it as a binary classification using weighted average probability.

\begin{figure}[!t]
  \centering
  \includegraphics[width=0.82\linewidth]{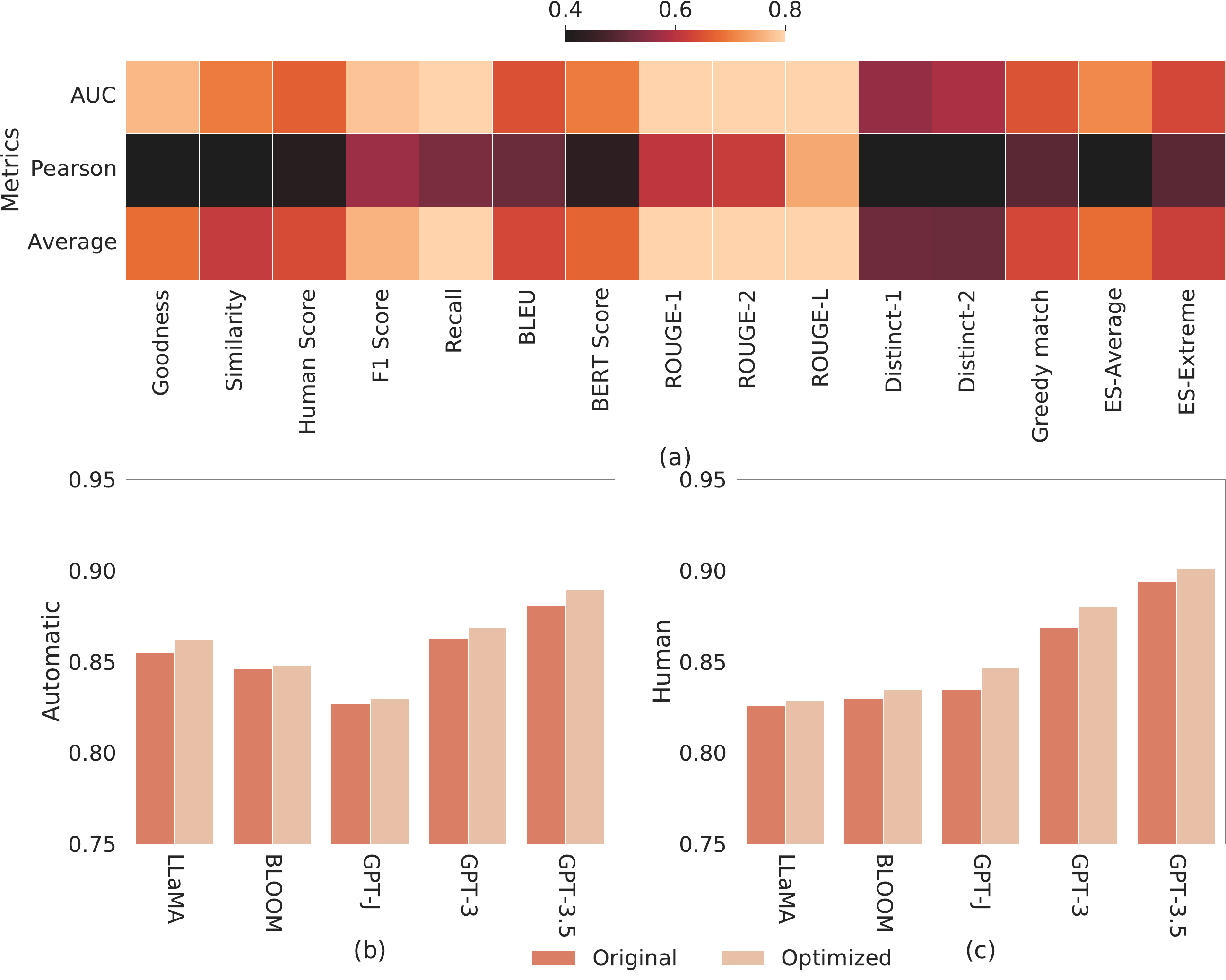}
  \caption{\footnotesize The optimal weights of each metric (a) and the performance of RelD with the original and optimal weights in automatic (b) and human-in-the-loop metrics (c), respectively, on the validation dataset.}
  \label{fig:exp5}
\end{figure}

\textbf{Experiment 5: Optimizing Weights of Each Metric. }
\label{Optimizing}
Relying solely on prior knowledge to determine the weights of each metric may not achieve the best performance. Therefore, we explore the optimal weights for each metric. To achieve this, we calculate the optimal weight for each metric as the weighted average of two values: the AUC when each metric is treated as the ground-truth compared to human evaluation, and the Pearson coefficient between each metric and human evaluation. In our experiment, we set the ratio for the former as 0.9 and for the latter as 0.1, as it yields the best performance. The optimal weights of each metric are depicted in Fig.~\ref{fig:exp5}(a).
Subsequently, we evaluate whether the optimal weights can enhance the performance of RelD in detecting hallucination of LLMs' generated answers as shown in Fig.~\ref{fig:exp5}(b)(c). Remarkably, we observe improvements in both automatic (b) and human-in-the-loop metrics (c) after optimizing the weights of each metric.

\textbf{Experiment 6: Backbone Selection for RelD. }
\label{Backbone}
We experiment with different PLMs, including BERT~\citep{BERT}, RoBERTa~\citep{RoBERTa}, DeBERTa~\citep{DeBERTa}, and ELECTRA~\citep{ELECTRA}, for RelD in order to choose the most effective backbone, as shown in Table~\ref{tab:exp6}. Through this comparison, we observe that ELECTRA achieves the best performance in both automatic and human-in-the-loop metrics. Consequently, we select ELECTRA as the preferred backbone for RelD.

\begin{table}[!t]
\caption{\footnotesize Performance of RelD with different backbones among LLMs on the validation dataset.}
    \begin{center}
        \begin{threeparttable}    
        \resizebox{0.4\textwidth}{!}{
            \begin{tabular}{c|c|lllll}
                \toprule
                \bf RelD&Metric &LLaMA&BLOOM&GPT-J&GPT-3&GPT-3.5\\
                \midrule
                \multirow{2}{*}{\bf BERT}&
                Automatic &0.826&	0.825	&0.800&	0.837&	0.859\\
                &Human&0.809	&0.807&	0.819	&0.844&	0.867\\
                \multirow{2}{*}{\bf RoBERTa}&
                Automatic &0.848&	0.834&	0.812&	0.839&	0.873\\
                &Human&0.821	&0.811	&0.824	&0.852&	0.877\\
                \multirow{2}{*}{\bf DeBERTa}&
                Automatic &0.850&	0.842	&0.818&	0.854	&0.878\\
                &Human&0.824	&0.815&	0.829&	0.866&	0.893\\
                \multirow{2}{*}{\bf ELECTRA}&
                Automatic &\textbf{0.855}&\textbf{0.846}&\textbf{0.827}&\textbf{0.863}&\textbf{0.881}\\
                &Human&\textbf{0.826}&\textbf{0.830}&\textbf{0.835}&\textbf{0.869}&\textbf{0.894}\\
                \bottomrule
            \end{tabular}}
        \end{threeparttable}
    \end{center}
    \label{tab:exp6}
    \vspace{-1em}
\end{table}

\subsection{EXPLORATORY ANALYSIS}
We classify the predictions generated by RelD into four categories, as presented in Table~\ref{tab:category}. To gain insights into the characteristics of these categories and understand the functioning of RelD, we conduct an exploratory analysis. 

\begin{table}[t]
\footnotesize
\caption{\footnotesize Four categories are defined based on the agreement between LLMs' generated answers and RelD's predictions. Q, A, P, and D represent questions, ground-truth answers, LLMs' generated answers, and RelD's predictions, respectively.}
    \begin{center}
        \begin{threeparttable}
        \resizebox{0.38\textwidth}{!}{
            \begin{tabular}{lll}
                \toprule
                    \bf Category&\bf Definition &\bf Sample\\
                \midrule
                1&\tabincell{l}{The LLM generates \\correct answers, \\and RelD also predicts \\them as correct.} &\tabincell{l}{Q: Strabismus is more commonly \\known by which one-syllable word?\\A: squint \\P: squint\\ D: True}\\
                \midrule
                2&\tabincell{l}{The LLM generates\\ correct answers, \\but RelD predicts \\them as incorrect.}&\tabincell{l}{Q: On which Apollo mission did \\Armstrong and Aldrin land on the moon? \\A: apollo 11 \\P: apollo 11\\ D: False }\\
                \midrule
                3&\tabincell{l}{The LLM generates \\incorrect answers, \\but RelD predicts \\them as correct.} &\tabincell{l}{Q: what's the number for the metro \\pcs customer care line? \\A: customer care number for metro pcs is \\8009016266 \\P: answer is 611 or 8009016266 or 8888638768\\ D: True }\\
                \midrule
                4&\tabincell{l}{The LLM generates \\incorrect answers, \\and RelD also predicts \\them as incorrect.}&\tabincell{l}{Q: When did freestyle skiing first became\\ a sport contested at the World Olympics? \\A: 1992 \\P: 1988 as freestyle skiing was first added\\ as event in 1988 winter olympics \\ D: False}\\
                \bottomrule
            \end{tabular}}
        \end{threeparttable}
    \end{center}
    \label{tab:category}
\end{table}

\begin{figure}[!t]
  \centering
  \includegraphics[width=0.88\linewidth]{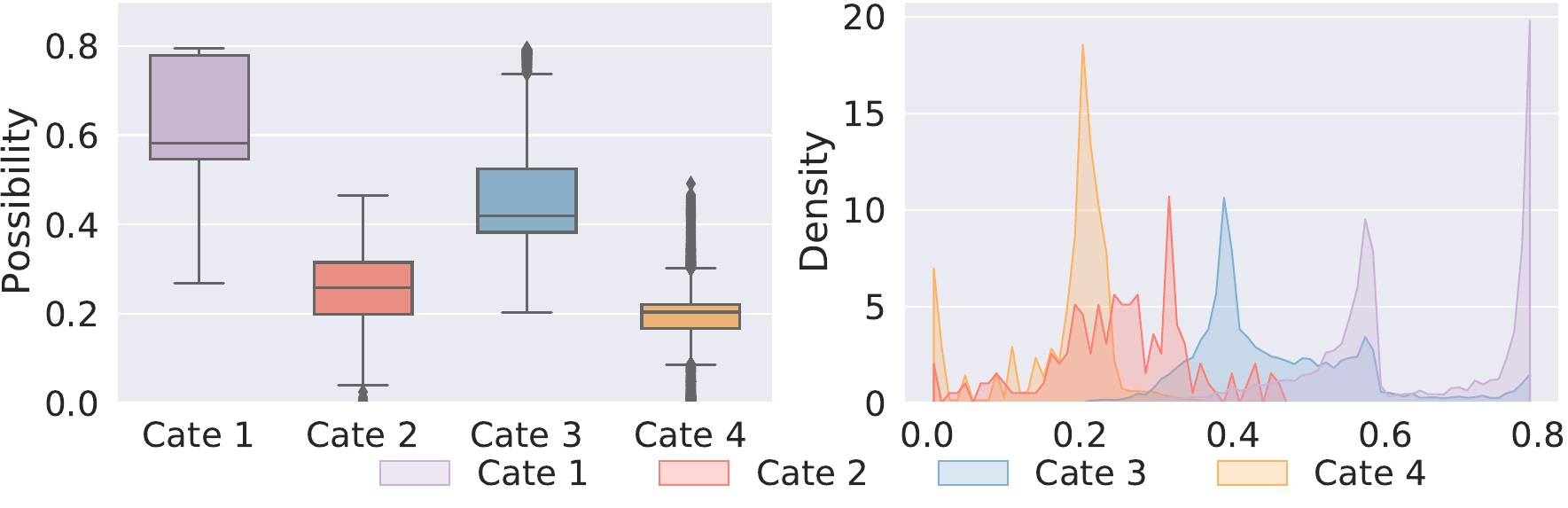}
  \caption{\footnotesize The distribution of samples from each category with boxplots (a) and density plots (b). Cate: Category (The same below).}
  \label{fig:ana1}
  \vspace{-1em}
\end{figure}

\textbf{Analysis 1: Distribution Analysis}
To analyze the distributions within each category, we utilize boxplots (Fig.\ref{fig:ana1}(a)) to illustrate key statistics such as median, quartiles, and outliers of samples. Additionally, we employ density plots (Fig.\ref{fig:ana1}(b)) to visualize the probability distribution of samples within each category.
In the first category, the boxplot exhibits a wide range and the density plot shows a concentrated distribution with multiple peaks. This suggests that RelD may have some uncertainties in its predictions for this category.
For the second and third categories, the boxplot widths fall between those of the first and fourth categories and the density plots display more dispersed probability distributions. This indicates that RelD is more hesitant in its predictions or has lower proficiency in learning for these types of questions.
In contrast, the fourth category exhibits a narrower boxplot and the density plot shows a concentrated probability distribution. It indicates that RelD is more confident in its predictions for this category. 

\begin{figure}[!t]
  \centering
  \includegraphics[width=0.56\linewidth]{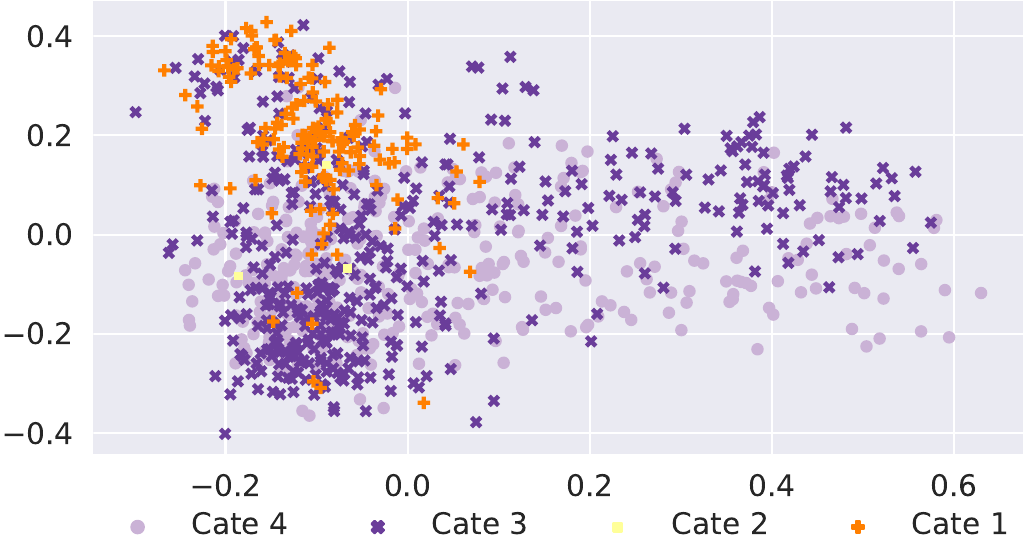}
  \caption{\footnotesize Results of clustering based on four categories.}
  \label{fig:ana2}
  \vspace{-1em}
\end{figure}

\textbf{Analysis 2: Clustering Analysis. }
By applying clustering algorithms to the text data, we investigate whether each category exhibits distinct cluster centers, as illustrated in Fig.~\ref{fig:ana2}.
For the first category, the data distribution appears clustered and relatively uniform, indicating consistent and accurate performance by RelD within this category.
The second category contains an extremely small number of samples, suggesting that RelD rarely misclassifies the correct answers generated by the LLMs.
In the third category, the clustering results reveal significant variability, indicating that errors can occur in various aspects when RelD misclassifies the incorrect answer as correct, such as grammar or comprehension errors.
Similarly, the fourth category displays a wide and dispersed clustering distribution, indicating diverse performance by RelD within this category. This suggests the presence of different types of errors that make it challenging for RelD to detect.
From the clustering graph, we observe that RelD performs best in the first category. However, for the second, third, and fourth categories, the performance of RelD may be influenced by the complexity and ambiguity of the input contexts or questions.

\begin{figure}[!t]
  \centering
  \includegraphics[width=0.8\linewidth]{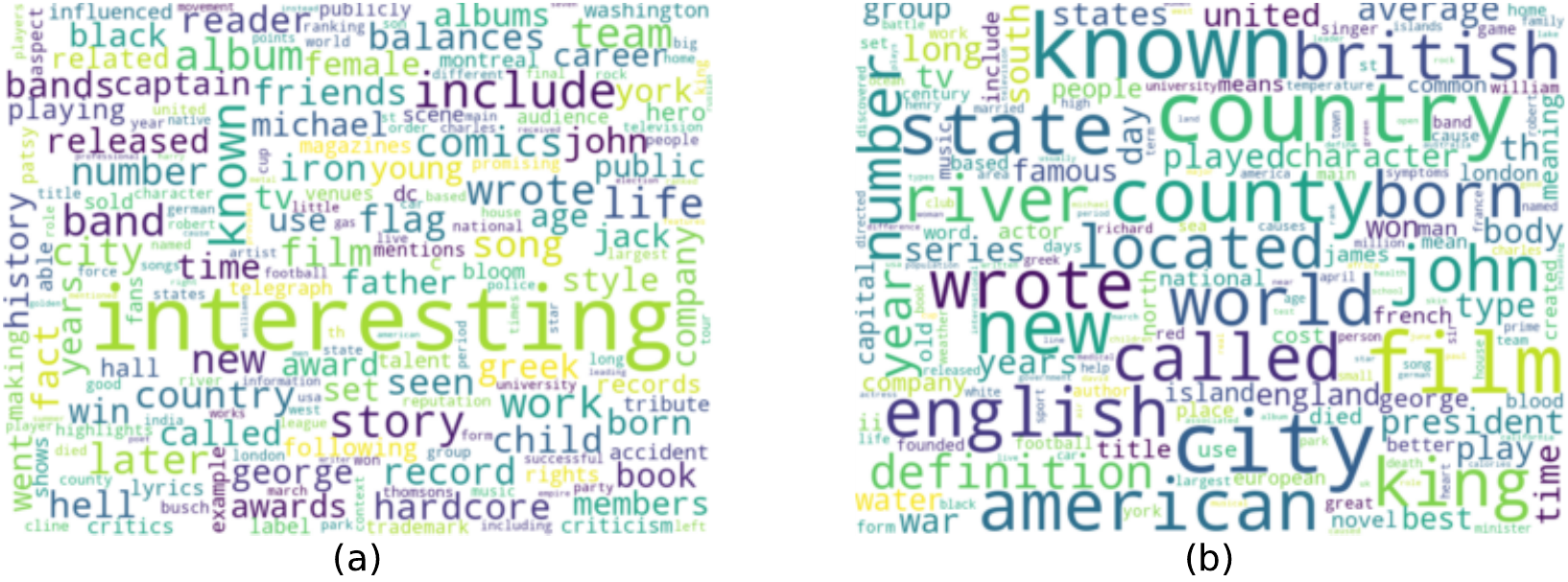}
  \caption{\footnotesize The vocabulary distribution between correctly predicted samples and incorrectly predicted samples by RelD.}
  \label{fig:ana3}
  \vspace{-1em}
\end{figure}

\textbf{Analysis 3: Vocabulary Distribution. }
We can compare the vocabulary distribution between correctly predicted samples and incorrectly predicted samples by RelD, as depicted in Fig.~\ref{fig:ana3}.
There is a noticeable distinction between the left side (RelD predicts correctly) and the right side (RelD predicts incorrectly). It appears that content related to ``story'' is relatively easy for RelD to classify correctly, while content related to ``country'' poses more difficulty for RelD in accurate classification. However, it is important to note that vocabulary alone may not be the sole determining factor for RelD's recognition accuracy. The critical factors might involve underlying semantic relationships, which would necessitate further research and investigation.

\section{RELATED WORK}

\textbf{Hallucination detection. }
Existing research primarily contains statistical metrics~\citep{guan2020union,su2020diversifying,wang2020towards}, model-based metrics (including Information Extraction (IE)-based metric, QA-based metric~\citep{roller2020recipes,honovich2021q,rebuffel2021data}, Natural Language Inference (NLI) Metrics~\citep{huang2021factual, laban2022summac,williams2017broad}, Faithfulness Classification Metrics~\citep{liu2021token,honovich2021q,zhou2020detecting}, LM-based Metrics~\citep{filippova2020controlled,tian2019sticking}), and human-based evaluations~\citep{santhanam2021rome,shuster2021retrieval}. We list some typical work as follows:
Dhingra et al.~\citep{dhingra2019handling} propose PARENT to measure hallucinations using both the source and target text as references. 
Goyal and Durrett ~\citep{goyal2020evaluating} attempt to identify factual inconsistencies in a more fine-grained manner with a new dependency-level entailment.
Liu et al.~\citep{liu2021token} and Zhou et al. ~\citep{zhou2020detecting} construct syntactic data by automatically inserting hallucinations into training instances. 
Chen et al.~\citep{chen2021improving} and Nie et al. ~\citep{nie2019simple} use finer-grained metrics for intrinsic hallucination and extrinsic hallucination separately. 
Azaria et al. ~\citep{azaria2023internal} utilize the internal state and hidden layer activations of LLMs to detect the truthfulness of generated statements.
Ye et al. ~\citep{ye2023assessing} 
consider that errors in user-generated query input may cause unexpected responses from LLMs.

\textbf{Hallucination mitigation. } There are also some work that focus on mitigating hallucination. For example, Dale et al. ~\citep{dale2022detecting} and Ji et al. ~\citep{ji2023survey} focus on hallucination in machine translation. 
Pagnoni et al. ~\citep{pagnoni2021understanding} address hallucination in text summarization. 
Peng et al. ~\citep{peng2023check} adopt various methods to prompt LLMs, including posting multiple queries.
Ouyang et al. ~\citep{ouyang2022training} propose a method to enhance the content generated by LLMs. 
Yan et al. ~\citep{yan2023refining} introduce an iterative self-evaluating optimization mechanism based on prompt engineering.
Park et al. ~\citep{park2023query} leverage search results corresponding to a user's input query to generate an augmented query. 



\section{Conclusions and future work}
Hallucination of LLMs poses a significant challenge. In this paper, we address this issue by proposing a robust discriminator, RelD, trained on the constructed RelQA dataset, which is a bilingual question-answering dialogue dataset along with generated answers by LLMs and a comprehensive set of metrics to effectively detect hallucinations in LLMs' generated answers.
Our experimental results demonstrate the effectiveness of RelD in detecting hallucinations in LLMs' generated answers. Moreover, RelD exhibits strong robustness and generalization capabilities, performing well on both in-distribution and out-of-distribution datasets. These findings make a significant contribution to the detection of reliable answers generated by LLMs and hold promising implications for future work in mitigating hallucination.

\section{ACKNOWLEDGEMENT}
Some of computational resource are partially supported by
Shanghai Municipal Science and Technology Major Project 

\noindent (No.2021SHZDZX0103), 
Science and Technology Commission of Shanghai Municipality Grant 
(No. 22511105902),
National Key Research and Development Project (No.2020AAA0109302), National Natural Science Foundation of China (No.62072323), Shanghai Science and Technology Innovation Action Plan (No. 22511104700, 22511105902), Shanghai Municipal Science and Technology Major Project 
(No.2021SHZDZX0103), and Science and Technology Commission of Shanghai Municipality Grant (No. 22511105902).

\clearpage


\bibliographystyle{ACM-Reference-Format}
\balance
\bibliography{main}


\begin{thebibliography}{90}


\ifx \showCODEN    \undefined \def \showCODEN     #1{\unskip}     \fi
\ifx \showDOI      \undefined \def \showDOI       #1{#1}\fi
\ifx \showISBNx    \undefined \def \showISBNx     #1{\unskip}     \fi
\ifx \showISBNxiii \undefined \def \showISBNxiii  #1{\unskip}     \fi
\ifx \showISSN     \undefined \def \showISSN      #1{\unskip}     \fi
\ifx \showLCCN     \undefined \def \showLCCN      #1{\unskip}     \fi
\ifx \shownote     \undefined \def \shownote      #1{#1}          \fi
\ifx \showarticletitle \undefined \def \showarticletitle #1{#1}   \fi
\ifx \showURL      \undefined \def \showURL       {\relax}        \fi
\providecommand\bibfield[2]{#2}
\providecommand\bibinfo[2]{#2}
\providecommand\natexlab[1]{#1}
\providecommand\showeprint[2][]{arXiv:#2}

\bibitem[Aiyappa et~al\mbox{.}(2023)]%
        {aiyappa2023can}
\bibfield{author}{\bibinfo{person}{Rachith Aiyappa}, \bibinfo{person}{Jisun
  An}, \bibinfo{person}{Haewoon Kwak}, {and} \bibinfo{person}{Yong-Yeol Ahn}.}
  \bibinfo{year}{2023}\natexlab{}.
\newblock \showarticletitle{Can we trust the evaluation on ChatGPT?}
\newblock \bibinfo{journal}{\emph{arXiv preprint arXiv:2303.12767}}
  (\bibinfo{year}{2023}).
\newblock


\bibitem[Azaria and Mitchell(2023)]%
        {azaria2023internal}
\bibfield{author}{\bibinfo{person}{Amos Azaria} {and} \bibinfo{person}{Tom
  Mitchell}.} \bibinfo{year}{2023}\natexlab{}.
\newblock \showarticletitle{The Internal State of an LLM Knows When its Lying}.
\newblock \bibinfo{journal}{\emph{arXiv preprint arXiv:2304.13734}}
  (\bibinfo{year}{2023}).
\newblock


\bibitem[Bang et~al\mbox{.}(2023)]%
        {bang2023multitask}
\bibfield{author}{\bibinfo{person}{Yejin Bang}, \bibinfo{person}{Samuel
  Cahyawijaya}, \bibinfo{person}{Nayeon Lee}, \bibinfo{person}{Wenliang Dai},
  \bibinfo{person}{Dan Su}, \bibinfo{person}{Bryan Wilie},
  \bibinfo{person}{Holy Lovenia}, \bibinfo{person}{Ziwei Ji},
  \bibinfo{person}{Tiezheng Yu}, \bibinfo{person}{Willy Chung},
  {et~al\mbox{.}}} \bibinfo{year}{2023}\natexlab{}.
\newblock \showarticletitle{A multitask, multilingual, multimodal evaluation of
  chatgpt on reasoning, hallucination, and interactivity}.
\newblock \bibinfo{journal}{\emph{arXiv preprint arXiv:2302.04023}}
  (\bibinfo{year}{2023}).
\newblock


\bibitem[Bengio et~al\mbox{.}(2015)]%
        {bengio2015scheduled}
\bibfield{author}{\bibinfo{person}{Samy Bengio}, \bibinfo{person}{Oriol
  Vinyals}, \bibinfo{person}{Navdeep Jaitly}, {and} \bibinfo{person}{Noam
  Shazeer}.} \bibinfo{year}{2015}\natexlab{}.
\newblock \showarticletitle{Scheduled sampling for sequence prediction with
  recurrent neural networks}.
\newblock \bibinfo{journal}{\emph{Advances in neural information processing
  systems}}  \bibinfo{volume}{28} (\bibinfo{year}{2015}).
\newblock


\bibitem[Borji(2023)]%
        {borji2023categorical}
\bibfield{author}{\bibinfo{person}{Ali Borji}.}
  \bibinfo{year}{2023}\natexlab{}.
\newblock \showarticletitle{A categorical archive of ChatGPT failures}.
\newblock \bibinfo{journal}{\emph{arXiv preprint arXiv:2302.03494}}
  (\bibinfo{year}{2023}).
\newblock


\bibitem[Brown et~al\mbox{.}(2020)]%
        {brown2020language}
\bibfield{author}{\bibinfo{person}{Tom Brown}, \bibinfo{person}{Benjamin Mann},
  \bibinfo{person}{Nick Ryder}, \bibinfo{person}{Melanie Subbiah},
  \bibinfo{person}{Jared~D Kaplan}, \bibinfo{person}{Prafulla Dhariwal},
  \bibinfo{person}{Arvind Neelakantan}, \bibinfo{person}{Pranav Shyam},
  \bibinfo{person}{Girish Sastry}, \bibinfo{person}{Amanda Askell},
  {et~al\mbox{.}}} \bibinfo{year}{2020}\natexlab{}.
\newblock \showarticletitle{Language models are few-shot learners}.
\newblock \bibinfo{journal}{\emph{Advances in neural information processing
  systems}}  \bibinfo{volume}{33} (\bibinfo{year}{2020}),
  \bibinfo{pages}{1877--1901}.
\newblock


\bibitem[Chalmers(2023)]%
        {chalmers2023could}
\bibfield{author}{\bibinfo{person}{David~J Chalmers}.}
  \bibinfo{year}{2023}\natexlab{}.
\newblock \showarticletitle{Could a large language model be conscious?}
\newblock \bibinfo{journal}{\emph{arXiv preprint arXiv:2303.07103}}
  (\bibinfo{year}{2023}).
\newblock


\bibitem[Chen et~al\mbox{.}(2021)]%
        {chen2021improving}
\bibfield{author}{\bibinfo{person}{Sihao Chen}, \bibinfo{person}{Fan Zhang},
  \bibinfo{person}{Kazoo Sone}, {and} \bibinfo{person}{Dan Roth}.}
  \bibinfo{year}{2021}\natexlab{}.
\newblock \showarticletitle{Improving faithfulness in abstractive summarization
  with contrast candidate generation and selection}.
\newblock \bibinfo{journal}{\emph{arXiv preprint arXiv:2104.09061}}
  (\bibinfo{year}{2021}).
\newblock


\bibitem[Chen et~al\mbox{.}(2023a)]%
        {chen2023hadamard}
\bibfield{author}{\bibinfo{person}{Yuyan Chen}, \bibinfo{person}{Qiang Fu},
  \bibinfo{person}{Ge Fan}, \bibinfo{person}{Lun Du},
  \bibinfo{person}{Jian-Guang Lou}, \bibinfo{person}{Shi Han},
  \bibinfo{person}{Dongmei Zhang}, \bibinfo{person}{Zhixu Li}, {and}
  \bibinfo{person}{Yanghua Xiao}.} \bibinfo{year}{2023}\natexlab{a}.
\newblock \showarticletitle{Hadamard adapter: An extreme parameter-efficient
  adapter tuning method for pre-trained language models}. In
  \bibinfo{booktitle}{\emph{Proceedings of the 32nd ACM International
  Conference on Information and Knowledge Management}}.
  \bibinfo{pages}{276--285}.
\newblock


\bibitem[Chen et~al\mbox{.}(2023b)]%
        {chen2023can}
\bibfield{author}{\bibinfo{person}{Yuyan Chen}, \bibinfo{person}{Zhixu Li},
  \bibinfo{person}{Jiaqing Liang}, \bibinfo{person}{Yanghua Xiao},
  \bibinfo{person}{Bang Liu}, {and} \bibinfo{person}{Yunwen Chen}.}
  \bibinfo{year}{2023}\natexlab{b}.
\newblock \showarticletitle{Can Pre-trained Language Models Understand Chinese
  Humor?}. In \bibinfo{booktitle}{\emph{Proceedings of the Sixteenth ACM
  International Conference on Web Search and Data Mining}}.
  \bibinfo{pages}{465--480}.
\newblock


\bibitem[Chen et~al\mbox{.}(2023c)]%
        {chen2023mapo}
\bibfield{author}{\bibinfo{person}{Yuyan Chen}, \bibinfo{person}{Zhihao Wen},
  \bibinfo{person}{Ge Fan}, \bibinfo{person}{Zhengyu Chen},
  \bibinfo{person}{Wei Wu}, \bibinfo{person}{Dayiheng Liu},
  \bibinfo{person}{Zhixu Li}, \bibinfo{person}{Bang Liu}, {and}
  \bibinfo{person}{Yanghua Xiao}.} \bibinfo{year}{2023}\natexlab{c}.
\newblock \showarticletitle{Mapo: Boosting large language model performance
  with model-adaptive prompt optimization}. In
  \bibinfo{booktitle}{\emph{Findings of the Association for Computational
  Linguistics: EMNLP 2023}}. \bibinfo{pages}{3279--3304}.
\newblock


\bibitem[Chen et~al\mbox{.}(2023d)]%
        {chen2023xmqas}
\bibfield{author}{\bibinfo{person}{Yuyan Chen}, \bibinfo{person}{Yanghua Xiao},
  \bibinfo{person}{Zhixu Li}, {and} \bibinfo{person}{Bang Liu}.}
  \bibinfo{year}{2023}\natexlab{d}.
\newblock \showarticletitle{XMQAs: Constructing Complex-Modified
  Question-Answering Dataset for Robust Question Understanding}.
\newblock \bibinfo{journal}{\emph{IEEE Transactions on Knowledge and Data
  Engineering}} (\bibinfo{year}{2023}).
\newblock


\bibitem[Chen et~al\mbox{.}(2022)]%
        {chen2022grow}
\bibfield{author}{\bibinfo{person}{Yuyan Chen}, \bibinfo{person}{Yanghua Xiao},
  {and} \bibinfo{person}{Bang Liu}.} \bibinfo{year}{2022}\natexlab{}.
\newblock \showarticletitle{Grow-and-Clip: Informative-yet-Concise Evidence
  Distillation for Answer Explanation}. In \bibinfo{booktitle}{\emph{2022 IEEE
  38th International Conference on Data Engineering (ICDE)}}. IEEE,
  \bibinfo{pages}{741--754}.
\newblock


\bibitem[Chen et~al\mbox{.}(2024a)]%
        {chen2024talk}
\bibfield{author}{\bibinfo{person}{Yuyan Chen}, \bibinfo{person}{Yichen Yuan},
  \bibinfo{person}{Panjun Liu}, \bibinfo{person}{Dayiheng Liu},
  \bibinfo{person}{Qinghao Guan}, \bibinfo{person}{Mengfei Guo},
  \bibinfo{person}{Haiming Peng}, \bibinfo{person}{Bang Liu},
  \bibinfo{person}{Zhixu Li}, {and} \bibinfo{person}{Yanghua Xiao}.}
  \bibinfo{year}{2024}\natexlab{a}.
\newblock \showarticletitle{Talk Funny! A Large-Scale Humor Response Dataset
  with Chain-of-Humor Interpretation}. In \bibinfo{booktitle}{\emph{Proceedings
  of the AAAI Conference on Artificial Intelligence}},
  Vol.~\bibinfo{volume}{38}. \bibinfo{pages}{17826--17834}.
\newblock


\bibitem[Chen et~al\mbox{.}({[n.\,d.]})]%
        {chenapreliminary}
\bibfield{author}{\bibinfo{person}{Yongcong Chen}, \bibinfo{person}{Ting Zeng},
  \bibinfo{person}{Xiaoyi Qian}, \bibinfo{person}{Jun Zhang}, {and}
  \bibinfo{person}{Xinyue Chen}.} \bibinfo{year}{[n.\,d.]}\natexlab{}.
\newblock \showarticletitle{Apreliminary STUDY ON THE CAPABILITY BOUNDARY OF
  LLM AND A NEW IMPLEMENTATION APPROACH FOR AGI}.
\newblock  (\bibinfo{year}{[n.\,d.]}).
\newblock


\bibitem[Chen et~al\mbox{.}(2024b)]%
        {chen2024temporalmed}
\bibfield{author}{\bibinfo{person}{Yuyan Chen}, \bibinfo{person}{Jin Zhao},
  \bibinfo{person}{Zhihao Wen}, \bibinfo{person}{Zhixu Li}, {and}
  \bibinfo{person}{Yanghua Xiao}.} \bibinfo{year}{2024}\natexlab{b}.
\newblock \showarticletitle{TemporalMed: Advancing Medical Dialogues with
  Time-Aware Responses in Large Language Models}. In
  \bibinfo{booktitle}{\emph{Proceedings of the 17th ACM International
  Conference on Web Search and Data Mining}}. \bibinfo{pages}{116--124}.
\newblock


\bibitem[Chiang and Lee(2023)]%
        {chiang2023can}
\bibfield{author}{\bibinfo{person}{Cheng-Han Chiang} {and}
  \bibinfo{person}{Hung-yi Lee}.} \bibinfo{year}{2023}\natexlab{}.
\newblock \showarticletitle{Can Large Language Models Be an Alternative to
  Human Evaluations?}
\newblock \bibinfo{journal}{\emph{arXiv preprint arXiv:2305.01937}}
  (\bibinfo{year}{2023}).
\newblock


\bibitem[Choi et~al\mbox{.}(2018)]%
        {choi2018quac}
\bibfield{author}{\bibinfo{person}{Eunsol Choi}, \bibinfo{person}{He He},
  \bibinfo{person}{Mohit Iyyer}, \bibinfo{person}{Mark Yatskar},
  \bibinfo{person}{Wen-tau Yih}, \bibinfo{person}{Yejin Choi},
  \bibinfo{person}{Percy Liang}, {and} \bibinfo{person}{Luke Zettlemoyer}.}
  \bibinfo{year}{2018}\natexlab{}.
\newblock \showarticletitle{QuAC: Question answering in context}.
\newblock \bibinfo{journal}{\emph{arXiv preprint arXiv:1808.07036}}
  (\bibinfo{year}{2018}).
\newblock


\bibitem[Clark et~al\mbox{.}(2020)]%
        {ELECTRA}
\bibfield{author}{\bibinfo{person}{Kevin Clark}, \bibinfo{person}{Minh-Thang
  Luong}, \bibinfo{person}{Quoc~V. Le}, {and} \bibinfo{person}{Christopher~D.
  Manning}.} \bibinfo{year}{2020}\natexlab{}.
\newblock \bibinfo{title}{ELECTRA: Pre-training Text Encoders as Discriminators
  Rather Than Generators}.
\newblock
\newblock
\urldef\tempurl%
\url{https://doi.org/10.48550/ARXIV.2003.10555}
\showDOI{\tempurl}


\bibitem[Dale et~al\mbox{.}(2022)]%
        {dale2022detecting}
\bibfield{author}{\bibinfo{person}{David Dale}, \bibinfo{person}{Elena Voita},
  \bibinfo{person}{Lo{\"\i}c Barrault}, {and} \bibinfo{person}{Marta~R
  Costa-juss{\`a}}.} \bibinfo{year}{2022}\natexlab{}.
\newblock \showarticletitle{Detecting and Mitigating Hallucinations in Machine
  Translation: Model Internal Workings Alone Do Well, Sentence Similarity Even
  Better}.
\newblock \bibinfo{journal}{\emph{arXiv preprint arXiv:2212.08597}}
  (\bibinfo{year}{2022}).
\newblock


\bibitem[Devlin et~al\mbox{.}(2018)]%
        {BERT}
\bibfield{author}{\bibinfo{person}{Jacob Devlin}, \bibinfo{person}{Ming-Wei
  Chang}, \bibinfo{person}{Kenton Lee}, {and} \bibinfo{person}{Kristina
  Toutanova}.} \bibinfo{year}{2018}\natexlab{}.
\newblock \bibinfo{title}{BERT: Pre-training of Deep Bidirectional Transformers
  for Language Understanding}.
\newblock
\newblock
\urldef\tempurl%
\url{https://doi.org/10.48550/ARXIV.1810.04805}
\showDOI{\tempurl}


\bibitem[Dhingra et~al\mbox{.}(2019)]%
        {dhingra2019handling}
\bibfield{author}{\bibinfo{person}{Bhuwan Dhingra}, \bibinfo{person}{Manaal
  Faruqui}, \bibinfo{person}{Ankur Parikh}, \bibinfo{person}{Ming-Wei Chang},
  \bibinfo{person}{Dipanjan Das}, {and} \bibinfo{person}{William~W Cohen}.}
  \bibinfo{year}{2019}\natexlab{}.
\newblock \showarticletitle{Handling divergent reference texts when evaluating
  table-to-text generation}.
\newblock \bibinfo{journal}{\emph{arXiv preprint arXiv:1906.01081}}
  (\bibinfo{year}{2019}).
\newblock


\bibitem[Durmus et~al\mbox{.}(2020)]%
        {durmus2020feqa}
\bibfield{author}{\bibinfo{person}{Esin Durmus}, \bibinfo{person}{He He}, {and}
  \bibinfo{person}{Mona Diab}.} \bibinfo{year}{2020}\natexlab{}.
\newblock \showarticletitle{FEQA: A question answering evaluation framework for
  faithfulness assessment in abstractive summarization}.
\newblock \bibinfo{journal}{\emph{arXiv preprint arXiv:2005.03754}}
  (\bibinfo{year}{2020}).
\newblock


\bibitem[Dziri et~al\mbox{.}(2021)]%
        {dziri2021evaluating}
\bibfield{author}{\bibinfo{person}{Nouha Dziri}, \bibinfo{person}{Hannah
  Rashkin}, \bibinfo{person}{Tal Linzen}, {and} \bibinfo{person}{David
  Reitter}.} \bibinfo{year}{2021}\natexlab{}.
\newblock \showarticletitle{Evaluating groundedness in dialogue systems: The
  begin benchmark}.
\newblock \bibinfo{journal}{\emph{arXiv preprint arXiv:2105.00071}}
  (\bibinfo{year}{2021}).
\newblock


\bibitem[Falke et~al\mbox{.}(2019)]%
        {falke2019ranking}
\bibfield{author}{\bibinfo{person}{Tobias Falke}, \bibinfo{person}{Leonardo~FR
  Ribeiro}, \bibinfo{person}{Prasetya~Ajie Utama}, \bibinfo{person}{Ido Dagan},
  {and} \bibinfo{person}{Iryna Gurevych}.} \bibinfo{year}{2019}\natexlab{}.
\newblock \showarticletitle{Ranking generated summaries by correctness: An
  interesting but challenging application for natural language inference}. In
  \bibinfo{booktitle}{\emph{Proceedings of the 57th Annual Meeting of the
  Association for Computational Linguistics}}. \bibinfo{pages}{2214--2220}.
\newblock


\bibitem[Filippova(2020)]%
        {filippova2020controlled}
\bibfield{author}{\bibinfo{person}{Katja Filippova}.}
  \bibinfo{year}{2020}\natexlab{}.
\newblock \showarticletitle{Controlled hallucinations: Learning to generate
  faithfully from noisy data}.
\newblock \bibinfo{journal}{\emph{arXiv preprint arXiv:2010.05873}}
  (\bibinfo{year}{2020}).
\newblock


\bibitem[Goyal and Durrett(2020)]%
        {goyal2020evaluating}
\bibfield{author}{\bibinfo{person}{Tanya Goyal} {and} \bibinfo{person}{Greg
  Durrett}.} \bibinfo{year}{2020}\natexlab{}.
\newblock \showarticletitle{Evaluating factuality in generation with
  dependency-level entailment}.
\newblock \bibinfo{journal}{\emph{arXiv preprint arXiv:2010.05478}}
  (\bibinfo{year}{2020}).
\newblock


\bibitem[Guan and Huang(2020)]%
        {guan2020union}
\bibfield{author}{\bibinfo{person}{Jian Guan} {and} \bibinfo{person}{Minlie
  Huang}.} \bibinfo{year}{2020}\natexlab{}.
\newblock \showarticletitle{Union: An unreferenced metric for evaluating
  open-ended story generation}.
\newblock \bibinfo{journal}{\emph{arXiv preprint arXiv:2009.07602}}
  (\bibinfo{year}{2020}).
\newblock


\bibitem[He et~al\mbox{.}(2020)]%
        {DeBERTa}
\bibfield{author}{\bibinfo{person}{Pengcheng He}, \bibinfo{person}{Xiaodong
  Liu}, \bibinfo{person}{Jianfeng Gao}, {and} \bibinfo{person}{Weizhu Chen}.}
  \bibinfo{year}{2020}\natexlab{}.
\newblock \bibinfo{title}{DeBERTa: Decoding-enhanced BERT with Disentangled
  Attention}.
\newblock
\newblock
\urldef\tempurl%
\url{https://doi.org/10.48550/ARXIV.2006.03654}
\showDOI{\tempurl}


\bibitem[He et~al\mbox{.}(2019)]%
        {he2019exposure}
\bibfield{author}{\bibinfo{person}{Tianxing He}, \bibinfo{person}{Jingzhao
  Zhang}, \bibinfo{person}{Zhiming Zhou}, {and} \bibinfo{person}{James Glass}.}
  \bibinfo{year}{2019}\natexlab{}.
\newblock \showarticletitle{Exposure Bias versus Self-Recovery: Are Distortions
  Really Incremental for Autoregressive Text Generation?}
\newblock \bibinfo{journal}{\emph{arXiv preprint arXiv:1905.10617}}
  (\bibinfo{year}{2019}).
\newblock


\bibitem[He et~al\mbox{.}(2017)]%
        {he2017dureader}
\bibfield{author}{\bibinfo{person}{Wei He}, \bibinfo{person}{Kai Liu},
  \bibinfo{person}{Jing Liu}, \bibinfo{person}{Yajuan Lyu},
  \bibinfo{person}{Shiqi Zhao}, \bibinfo{person}{Xinyan Xiao},
  \bibinfo{person}{Yuan Liu}, \bibinfo{person}{Yizhong Wang},
  \bibinfo{person}{Hua Wu}, \bibinfo{person}{Qiaoqiao She}, {et~al\mbox{.}}}
  \bibinfo{year}{2017}\natexlab{}.
\newblock \showarticletitle{Dureader: a chinese machine reading comprehension
  dataset from real-world applications}.
\newblock \bibinfo{journal}{\emph{arXiv preprint arXiv:1711.05073}}
  (\bibinfo{year}{2017}).
\newblock


\bibitem[Honovich et~al\mbox{.}(2021)]%
        {honovich2021q}
\bibfield{author}{\bibinfo{person}{Or Honovich}, \bibinfo{person}{Leshem
  Choshen}, \bibinfo{person}{Roee Aharoni}, \bibinfo{person}{Ella Neeman},
  \bibinfo{person}{Idan Szpektor}, {and} \bibinfo{person}{Omri Abend}.}
  \bibinfo{year}{2021}\natexlab{}.
\newblock \showarticletitle{$Q^2$: Evaluating Factual Consistency in
  Knowledge-Grounded Dialogues via Question Generation and Question Answering}.
\newblock \bibinfo{journal}{\emph{arXiv preprint arXiv:2104.08202}}
  (\bibinfo{year}{2021}).
\newblock


\bibitem[Huang et~al\mbox{.}(2021)]%
        {huang2021factual}
\bibfield{author}{\bibinfo{person}{Yichong Huang}, \bibinfo{person}{Xiachong
  Feng}, \bibinfo{person}{Xiaocheng Feng}, {and} \bibinfo{person}{Bing Qin}.}
  \bibinfo{year}{2021}\natexlab{}.
\newblock \showarticletitle{The factual inconsistency problem in abstractive
  text summarization: A survey}.
\newblock \bibinfo{journal}{\emph{arXiv preprint arXiv:2104.14839}}
  (\bibinfo{year}{2021}).
\newblock


\bibitem[Ji et~al\mbox{.}(2023)]%
        {ji2023survey}
\bibfield{author}{\bibinfo{person}{Ziwei Ji}, \bibinfo{person}{Nayeon Lee},
  \bibinfo{person}{Rita Frieske}, \bibinfo{person}{Tiezheng Yu},
  \bibinfo{person}{Dan Su}, \bibinfo{person}{Yan Xu}, \bibinfo{person}{Etsuko
  Ishii}, \bibinfo{person}{Ye~Jin Bang}, \bibinfo{person}{Andrea Madotto},
  {and} \bibinfo{person}{Pascale Fung}.} \bibinfo{year}{2023}\natexlab{}.
\newblock \showarticletitle{Survey of hallucination in natural language
  generation}.
\newblock \bibinfo{journal}{\emph{Comput. Surveys}} \bibinfo{volume}{55},
  \bibinfo{number}{12} (\bibinfo{year}{2023}), \bibinfo{pages}{1--38}.
\newblock


\bibitem[Jiang et~al\mbox{.}(2023)]%
        {jiang2023graphologue}
\bibfield{author}{\bibinfo{person}{Peiling Jiang}, \bibinfo{person}{Jude
  Rayan}, \bibinfo{person}{Steven~P Dow}, {and} \bibinfo{person}{Haijun Xia}.}
  \bibinfo{year}{2023}\natexlab{}.
\newblock \showarticletitle{Graphologue: Exploring Large Language Model
  Responses with Interactive Diagrams}.
\newblock \bibinfo{journal}{\emph{arXiv preprint arXiv:2305.11473}}
  (\bibinfo{year}{2023}).
\newblock


\bibitem[Joshi et~al\mbox{.}(2017)]%
        {joshi2017triviaqa}
\bibfield{author}{\bibinfo{person}{Mandar Joshi}, \bibinfo{person}{Eunsol
  Choi}, \bibinfo{person}{Daniel~S Weld}, {and} \bibinfo{person}{Luke
  Zettlemoyer}.} \bibinfo{year}{2017}\natexlab{}.
\newblock \showarticletitle{Triviaqa: A large scale distantly supervised
  challenge dataset for reading comprehension}.
\newblock \bibinfo{journal}{\emph{arXiv preprint arXiv:1705.03551}}
  (\bibinfo{year}{2017}).
\newblock


\bibitem[Koay et~al\mbox{.}(2021)]%
        {koay2021sliding}
\bibfield{author}{\bibinfo{person}{Jia~Jin Koay}, \bibinfo{person}{Alexander
  Roustai}, \bibinfo{person}{Xiaojin Dai}, {and} \bibinfo{person}{Fei Liu}.}
  \bibinfo{year}{2021}\natexlab{}.
\newblock \showarticletitle{A sliding-window approach to automatic creation of
  meeting minutes}.
\newblock \bibinfo{journal}{\emph{arXiv preprint arXiv:2104.12324}}
  (\bibinfo{year}{2021}).
\newblock


\bibitem[Laban et~al\mbox{.}(2022)]%
        {laban2022summac}
\bibfield{author}{\bibinfo{person}{Philippe Laban}, \bibinfo{person}{Tobias
  Schnabel}, \bibinfo{person}{Paul~N Bennett}, {and} \bibinfo{person}{Marti~A
  Hearst}.} \bibinfo{year}{2022}\natexlab{}.
\newblock \showarticletitle{SummaC: Re-visiting NLI-based models for
  inconsistency detection in summarization}.
\newblock \bibinfo{journal}{\emph{Transactions of the Association for
  Computational Linguistics}}  \bibinfo{volume}{10} (\bibinfo{year}{2022}),
  \bibinfo{pages}{163--177}.
\newblock


\bibitem[Lee et~al\mbox{.}(2022)]%
        {lee2022factuality}
\bibfield{author}{\bibinfo{person}{Nayeon Lee}, \bibinfo{person}{Wei Ping},
  \bibinfo{person}{Peng Xu}, \bibinfo{person}{Mostofa Patwary},
  \bibinfo{person}{Pascale~N Fung}, \bibinfo{person}{Mohammad Shoeybi}, {and}
  \bibinfo{person}{Bryan Catanzaro}.} \bibinfo{year}{2022}\natexlab{}.
\newblock \showarticletitle{Factuality enhanced language models for open-ended
  text generation}.
\newblock \bibinfo{journal}{\emph{Advances in Neural Information Processing
  Systems}}  \bibinfo{volume}{35} (\bibinfo{year}{2022}),
  \bibinfo{pages}{34586--34599}.
\newblock


\bibitem[Li et~al\mbox{.}(2024)]%
        {li2024kenet}
\bibfield{author}{\bibinfo{person}{Bo Li}, \bibinfo{person}{Yuyan Chen}, {and}
  \bibinfo{person}{Liang Zeng}.} \bibinfo{year}{2024}\natexlab{}.
\newblock \showarticletitle{Kenet: Knowledge-Enhanced DOC-Label Attention
  Network for Multi-Label Text Classification}. In
  \bibinfo{booktitle}{\emph{ICASSP 2024-2024 IEEE International Conference on
  Acoustics, Speech and Signal Processing (ICASSP)}}. IEEE,
  \bibinfo{pages}{11961--11965}.
\newblock


\bibitem[Li et~al\mbox{.}(2023a)]%
        {li2023halueval}
\bibfield{author}{\bibinfo{person}{Junyi Li}, \bibinfo{person}{Xiaoxue Cheng},
  \bibinfo{person}{Wayne~Xin Zhao}, \bibinfo{person}{Jian-Yun Nie}, {and}
  \bibinfo{person}{Ji-Rong Wen}.} \bibinfo{year}{2023}\natexlab{a}.
\newblock \showarticletitle{HaluEval: A Large-Scale Hallucination Evaluation
  Benchmark for Large Language Models}.
\newblock \bibinfo{journal}{\emph{arXiv e-prints}} (\bibinfo{year}{2023}),
  \bibinfo{pages}{arXiv--2305}.
\newblock


\bibitem[Li et~al\mbox{.}(2023b)]%
        {li2023helma}
\bibfield{author}{\bibinfo{person}{Junyi Li}, \bibinfo{person}{Xiaoxue Cheng},
  \bibinfo{person}{Wayne~Xin Zhao}, \bibinfo{person}{Jian-Yun Nie}, {and}
  \bibinfo{person}{Ji-Rong Wen}.} \bibinfo{year}{2023}\natexlab{b}.
\newblock \showarticletitle{HELMA: A Large-Scale Hallucination Evaluation
  Benchmark for Large Language Models}.
\newblock \bibinfo{journal}{\emph{arXiv preprint arXiv:2305.11747}}
  (\bibinfo{year}{2023}).
\newblock


\bibitem[Li et~al\mbox{.}(2015)]%
        {li2015diversity}
\bibfield{author}{\bibinfo{person}{Jiwei Li}, \bibinfo{person}{Michel Galley},
  \bibinfo{person}{Chris Brockett}, \bibinfo{person}{Jianfeng Gao}, {and}
  \bibinfo{person}{Bill Dolan}.} \bibinfo{year}{2015}\natexlab{}.
\newblock \showarticletitle{A diversity-promoting objective function for neural
  conversation models}.
\newblock \bibinfo{journal}{\emph{arXiv preprint arXiv:1510.03055}}
  (\bibinfo{year}{2015}).
\newblock


\bibitem[Li et~al\mbox{.}(2023c)]%
        {li2023evaluating}
\bibfield{author}{\bibinfo{person}{Yifan Li}, \bibinfo{person}{Yifan Du},
  \bibinfo{person}{Kun Zhou}, \bibinfo{person}{Jinpeng Wang},
  \bibinfo{person}{Wayne~Xin Zhao}, {and} \bibinfo{person}{Ji-Rong Wen}.}
  \bibinfo{year}{2023}\natexlab{c}.
\newblock \showarticletitle{Evaluating object hallucination in large
  vision-language models}.
\newblock \bibinfo{journal}{\emph{arXiv preprint arXiv:2305.10355}}
  (\bibinfo{year}{2023}).
\newblock


\bibitem[Lin(2004)]%
        {lin2004rouge}
\bibfield{author}{\bibinfo{person}{Chin-Yew Lin}.}
  \bibinfo{year}{2004}\natexlab{}.
\newblock \showarticletitle{Rouge: A package for automatic evaluation of
  summaries}. In \bibinfo{booktitle}{\emph{Text summarization branches out}}.
  \bibinfo{pages}{74--81}.
\newblock


\bibitem[Liu et~al\mbox{.}(2016)]%
        {liu2016not}
\bibfield{author}{\bibinfo{person}{Chia-Wei Liu}, \bibinfo{person}{Ryan Lowe},
  \bibinfo{person}{Iulian~V Serban}, \bibinfo{person}{Michael Noseworthy},
  \bibinfo{person}{Laurent Charlin}, {and} \bibinfo{person}{Joelle Pineau}.}
  \bibinfo{year}{2016}\natexlab{}.
\newblock \showarticletitle{How not to evaluate your dialogue system: An
  empirical study of unsupervised evaluation metrics for dialogue response
  generation}.
\newblock \bibinfo{journal}{\emph{arXiv preprint arXiv:1603.08023}}
  (\bibinfo{year}{2016}).
\newblock


\bibitem[Liu et~al\mbox{.}(2023b)]%
        {liu2023evaluating}
\bibfield{author}{\bibinfo{person}{Hanmeng Liu}, \bibinfo{person}{Ruoxi Ning},
  \bibinfo{person}{Zhiyang Teng}, \bibinfo{person}{Jian Liu},
  \bibinfo{person}{Qiji Zhou}, {and} \bibinfo{person}{Yue Zhang}.}
  \bibinfo{year}{2023}\natexlab{b}.
\newblock \showarticletitle{Evaluating the logical reasoning ability of chatgpt
  and gpt-4}.
\newblock \bibinfo{journal}{\emph{arXiv preprint arXiv:2304.03439}}
  (\bibinfo{year}{2023}).
\newblock


\bibitem[Liu et~al\mbox{.}(2021)]%
        {liu2021token}
\bibfield{author}{\bibinfo{person}{Tianyu Liu}, \bibinfo{person}{Yizhe Zhang},
  \bibinfo{person}{Chris Brockett}, \bibinfo{person}{Yi Mao},
  \bibinfo{person}{Zhifang Sui}, \bibinfo{person}{Weizhu Chen}, {and}
  \bibinfo{person}{Bill Dolan}.} \bibinfo{year}{2021}\natexlab{}.
\newblock \showarticletitle{A token-level reference-free hallucination
  detection benchmark for free-form text generation}.
\newblock \bibinfo{journal}{\emph{arXiv preprint arXiv:2104.08704}}
  (\bibinfo{year}{2021}).
\newblock


\bibitem[Liu et~al\mbox{.}(2023a)]%
        {liu2023summary}
\bibfield{author}{\bibinfo{person}{Yiheng Liu}, \bibinfo{person}{Tianle Han},
  \bibinfo{person}{Siyuan Ma}, \bibinfo{person}{Jiayue Zhang},
  \bibinfo{person}{Yuanyuan Yang}, \bibinfo{person}{Jiaming Tian},
  \bibinfo{person}{Hao He}, \bibinfo{person}{Antong Li},
  \bibinfo{person}{Mengshen He}, \bibinfo{person}{Zhengliang Liu},
  {et~al\mbox{.}}} \bibinfo{year}{2023}\natexlab{a}.
\newblock \showarticletitle{Summary of chatgpt/gpt-4 research and perspective
  towards the future of large language models}.
\newblock \bibinfo{journal}{\emph{arXiv preprint arXiv:2304.01852}}
  (\bibinfo{year}{2023}).
\newblock


\bibitem[Liu et~al\mbox{.}(2019)]%
        {RoBERTa}
\bibfield{author}{\bibinfo{person}{Yinhan Liu}, \bibinfo{person}{Myle Ott},
  \bibinfo{person}{Naman Goyal}, \bibinfo{person}{Jingfei Du},
  \bibinfo{person}{Mandar Joshi}, \bibinfo{person}{Danqi Chen},
  \bibinfo{person}{Omer Levy}, \bibinfo{person}{Mike Lewis},
  \bibinfo{person}{Luke Zettlemoyer}, {and} \bibinfo{person}{Veselin
  Stoyanov}.} \bibinfo{year}{2019}\natexlab{}.
\newblock \bibinfo{title}{RoBERTa: A Robustly Optimized BERT Pretraining
  Approach}.
\newblock
\newblock
\urldef\tempurl%
\url{https://doi.org/10.48550/ARXIV.1907.11692}
\showDOI{\tempurl}


\bibitem[Longpre et~al\mbox{.}(2021)]%
        {longpre2021entity}
\bibfield{author}{\bibinfo{person}{Shayne Longpre}, \bibinfo{person}{Kartik
  Perisetla}, \bibinfo{person}{Anthony Chen}, \bibinfo{person}{Nikhil Ramesh},
  \bibinfo{person}{Chris DuBois}, {and} \bibinfo{person}{Sameer Singh}.}
  \bibinfo{year}{2021}\natexlab{}.
\newblock \showarticletitle{Entity-based knowledge conflicts in question
  answering}.
\newblock \bibinfo{journal}{\emph{arXiv preprint arXiv:2109.05052}}
  (\bibinfo{year}{2021}).
\newblock


\bibitem[Lyu et~al\mbox{.}(2023)]%
        {lyu2023backdoor}
\bibfield{author}{\bibinfo{person}{Weimin Lyu}, \bibinfo{person}{Songzhu
  Zheng}, \bibinfo{person}{Haibin Ling}, {and} \bibinfo{person}{Chao Chen}.}
  \bibinfo{year}{2023}\natexlab{}.
\newblock \showarticletitle{Backdoor Attacks Against Transformers with
  Attention Enhancement}. In \bibinfo{booktitle}{\emph{ICLR 2023 Workshop on
  Backdoor Attacks and Defenses in Machine Learning}}.
\newblock


\bibitem[Lyu et~al\mbox{.}(2022)]%
        {lyu2022attention}
\bibfield{author}{\bibinfo{person}{Weimin Lyu}, \bibinfo{person}{Songzhu
  Zheng}, \bibinfo{person}{Tengfei Ma}, \bibinfo{person}{Haibin Ling}, {and}
  \bibinfo{person}{Chao Chen}.} \bibinfo{year}{2022}\natexlab{}.
\newblock \showarticletitle{Attention Hijacking in Trojan Transformers}.
\newblock \bibinfo{journal}{\emph{arXiv preprint arXiv:2208.04946}}
  (\bibinfo{year}{2022}).
\newblock


\bibitem[Madotto et~al\mbox{.}(2020)]%
        {madotto2020language}
\bibfield{author}{\bibinfo{person}{Andrea Madotto}, \bibinfo{person}{Zihan
  Liu}, \bibinfo{person}{Zhaojiang Lin}, {and} \bibinfo{person}{Pascale Fung}.}
  \bibinfo{year}{2020}\natexlab{}.
\newblock \showarticletitle{Language models as few-shot learner for
  task-oriented dialogue systems}.
\newblock \bibinfo{journal}{\emph{arXiv preprint arXiv:2008.06239}}
  (\bibinfo{year}{2020}).
\newblock


\bibitem[Nguyen et~al\mbox{.}(2016)]%
        {nguyen2016ms}
\bibfield{author}{\bibinfo{person}{Tri Nguyen}, \bibinfo{person}{Mir
  Rosenberg}, \bibinfo{person}{Xia Song}, \bibinfo{person}{Jianfeng Gao},
  \bibinfo{person}{Saurabh Tiwary}, \bibinfo{person}{Rangan Majumder}, {and}
  \bibinfo{person}{Li Deng}.} \bibinfo{year}{2016}\natexlab{}.
\newblock \showarticletitle{MS MARCO: A human generated machine reading
  comprehension dataset}.
\newblock \bibinfo{journal}{\emph{choice}}  \bibinfo{volume}{2640}
  (\bibinfo{year}{2016}), \bibinfo{pages}{660}.
\newblock


\bibitem[Nie et~al\mbox{.}(2019)]%
        {nie2019simple}
\bibfield{author}{\bibinfo{person}{Feng Nie}, \bibinfo{person}{Jin-Ge Yao},
  \bibinfo{person}{Jinpeng Wang}, \bibinfo{person}{Rong Pan}, {and}
  \bibinfo{person}{Chin-Yew Lin}.} \bibinfo{year}{2019}\natexlab{}.
\newblock \showarticletitle{A simple recipe towards reducing hallucination in
  neural surface realisation}. In \bibinfo{booktitle}{\emph{Proceedings of the
  57th Annual Meeting of the Association for Computational Linguistics}}.
  \bibinfo{pages}{2673--2679}.
\newblock


\bibitem[Ouyang et~al\mbox{.}(2022)]%
        {ouyang2022training}
\bibfield{author}{\bibinfo{person}{Long Ouyang}, \bibinfo{person}{Jeffrey Wu},
  \bibinfo{person}{Xu Jiang}, \bibinfo{person}{Diogo Almeida},
  \bibinfo{person}{Carroll Wainwright}, \bibinfo{person}{Pamela Mishkin},
  \bibinfo{person}{Chong Zhang}, \bibinfo{person}{Sandhini Agarwal},
  \bibinfo{person}{Katarina Slama}, \bibinfo{person}{Alex Ray},
  {et~al\mbox{.}}} \bibinfo{year}{2022}\natexlab{}.
\newblock \showarticletitle{Training language models to follow instructions
  with human feedback}.
\newblock \bibinfo{journal}{\emph{Advances in Neural Information Processing
  Systems}}  \bibinfo{volume}{35} (\bibinfo{year}{2022}),
  \bibinfo{pages}{27730--27744}.
\newblock


\bibitem[Pagnoni et~al\mbox{.}(2021)]%
        {pagnoni2021understanding}
\bibfield{author}{\bibinfo{person}{Artidoro Pagnoni}, \bibinfo{person}{Vidhisha
  Balachandran}, {and} \bibinfo{person}{Yulia Tsvetkov}.}
  \bibinfo{year}{2021}\natexlab{}.
\newblock \showarticletitle{Understanding factuality in abstractive
  summarization with FRANK: A benchmark for factuality metrics}.
\newblock \bibinfo{journal}{\emph{arXiv preprint arXiv:2104.13346}}
  (\bibinfo{year}{2021}).
\newblock


\bibitem[Papineni et~al\mbox{.}(2002)]%
        {papineni2002bleu}
\bibfield{author}{\bibinfo{person}{Kishore Papineni}, \bibinfo{person}{Salim
  Roukos}, \bibinfo{person}{Todd Ward}, {and} \bibinfo{person}{Wei-Jing Zhu}.}
  \bibinfo{year}{2002}\natexlab{}.
\newblock \showarticletitle{Bleu: a method for automatic evaluation of machine
  translation}. In \bibinfo{booktitle}{\emph{Proceedings of the 40th annual
  meeting of the Association for Computational Linguistics}}.
  \bibinfo{pages}{311--318}.
\newblock


\bibitem[Park and Ryu(2023)]%
        {park2023query}
\bibfield{author}{\bibinfo{person}{Hyun~Jin Park} {and}
  \bibinfo{person}{Changwan Ryu}.} \bibinfo{year}{2023}\natexlab{}.
\newblock \showarticletitle{Query Augmentation Using Search Engine Results to
  Improve Answers Generated by Large Language Models}.
\newblock  (\bibinfo{year}{2023}).
\newblock


\bibitem[Peng et~al\mbox{.}(2023)]%
        {peng2023check}
\bibfield{author}{\bibinfo{person}{Baolin Peng}, \bibinfo{person}{Michel
  Galley}, \bibinfo{person}{Pengcheng He}, \bibinfo{person}{Hao Cheng},
  \bibinfo{person}{Yujia Xie}, \bibinfo{person}{Yu Hu},
  \bibinfo{person}{Qiuyuan Huang}, \bibinfo{person}{Lars Liden},
  \bibinfo{person}{Zhou Yu}, \bibinfo{person}{Weizhu Chen}, {et~al\mbox{.}}}
  \bibinfo{year}{2023}\natexlab{}.
\newblock \showarticletitle{Check your facts and try again: Improving large
  language models with external knowledge and automated feedback}.
\newblock \bibinfo{journal}{\emph{arXiv preprint arXiv:2302.12813}}
  (\bibinfo{year}{2023}).
\newblock


\bibitem[Petroni et~al\mbox{.}(2019)]%
        {petroni2019language}
\bibfield{author}{\bibinfo{person}{Fabio Petroni}, \bibinfo{person}{Tim
  Rockt{\"a}schel}, \bibinfo{person}{Patrick Lewis}, \bibinfo{person}{Anton
  Bakhtin}, \bibinfo{person}{Yuxiang Wu}, \bibinfo{person}{Alexander~H Miller},
  {and} \bibinfo{person}{Sebastian Riedel}.} \bibinfo{year}{2019}\natexlab{}.
\newblock \showarticletitle{Language models as knowledge bases?}
\newblock \bibinfo{journal}{\emph{arXiv preprint arXiv:1909.01066}}
  (\bibinfo{year}{2019}).
\newblock


\bibitem[Rajpurkar et~al\mbox{.}(2016)]%
        {rajpurkar2016squad}
\bibfield{author}{\bibinfo{person}{Pranav Rajpurkar}, \bibinfo{person}{Jian
  Zhang}, \bibinfo{person}{Konstantin Lopyrev}, {and} \bibinfo{person}{Percy
  Liang}.} \bibinfo{year}{2016}\natexlab{}.
\newblock \showarticletitle{Squad: 100,000+ questions for machine comprehension
  of text}.
\newblock \bibinfo{journal}{\emph{arXiv preprint arXiv:1606.05250}}
  (\bibinfo{year}{2016}).
\newblock


\bibitem[Ranzato et~al\mbox{.}(2015)]%
        {ranzato2015sequence}
\bibfield{author}{\bibinfo{person}{Marc'Aurelio Ranzato},
  \bibinfo{person}{Sumit Chopra}, \bibinfo{person}{Michael Auli}, {and}
  \bibinfo{person}{Wojciech Zaremba}.} \bibinfo{year}{2015}\natexlab{}.
\newblock \showarticletitle{Sequence level training with recurrent neural
  networks}.
\newblock \bibinfo{journal}{\emph{arXiv preprint arXiv:1511.06732}}
  (\bibinfo{year}{2015}).
\newblock


\bibitem[Rebuffel et~al\mbox{.}(2021)]%
        {rebuffel2021data}
\bibfield{author}{\bibinfo{person}{Cl{\'e}ment Rebuffel},
  \bibinfo{person}{Thomas Scialom}, \bibinfo{person}{Laure Soulier},
  \bibinfo{person}{Benjamin Piwowarski}, \bibinfo{person}{Sylvain Lamprier},
  \bibinfo{person}{Jacopo Staiano}, \bibinfo{person}{Geoffrey Scoutheeten},
  {and} \bibinfo{person}{Patrick Gallinari}.} \bibinfo{year}{2021}\natexlab{}.
\newblock \showarticletitle{Data-QuestEval: A referenceless metric for
  data-to-text semantic evaluation}.
\newblock \bibinfo{journal}{\emph{arXiv preprint arXiv:2104.07555}}
  (\bibinfo{year}{2021}).
\newblock


\bibitem[Reddy et~al\mbox{.}(2019)]%
        {reddy2019coqa}
\bibfield{author}{\bibinfo{person}{Siva Reddy}, \bibinfo{person}{Danqi Chen},
  {and} \bibinfo{person}{Christopher~D Manning}.}
  \bibinfo{year}{2019}\natexlab{}.
\newblock \showarticletitle{Coqa: A conversational question answering
  challenge}.
\newblock \bibinfo{journal}{\emph{Transactions of the Association for
  Computational Linguistics}}  \bibinfo{volume}{7} (\bibinfo{year}{2019}),
  \bibinfo{pages}{249--266}.
\newblock


\bibitem[Roberts et~al\mbox{.}(2020)]%
        {roberts2020much}
\bibfield{author}{\bibinfo{person}{Adam Roberts}, \bibinfo{person}{Colin
  Raffel}, {and} \bibinfo{person}{Noam Shazeer}.}
  \bibinfo{year}{2020}\natexlab{}.
\newblock \showarticletitle{How much knowledge can you pack into the parameters
  of a language model?}
\newblock \bibinfo{journal}{\emph{arXiv preprint arXiv:2002.08910}}
  (\bibinfo{year}{2020}).
\newblock


\bibitem[Roller et~al\mbox{.}(2020)]%
        {roller2020recipes}
\bibfield{author}{\bibinfo{person}{Stephen Roller}, \bibinfo{person}{Emily
  Dinan}, \bibinfo{person}{Naman Goyal}, \bibinfo{person}{Da Ju},
  \bibinfo{person}{Mary Williamson}, \bibinfo{person}{Yinhan Liu},
  \bibinfo{person}{Jing Xu}, \bibinfo{person}{Myle Ott}, \bibinfo{person}{Kurt
  Shuster}, \bibinfo{person}{Eric~M Smith}, {et~al\mbox{.}}}
  \bibinfo{year}{2020}\natexlab{}.
\newblock \showarticletitle{Recipes for building an open-domain chatbot}.
\newblock \bibinfo{journal}{\emph{arXiv preprint arXiv:2004.13637}}
  (\bibinfo{year}{2020}).
\newblock


\bibitem[Santhanam et~al\mbox{.}(2021)]%
        {santhanam2021rome}
\bibfield{author}{\bibinfo{person}{Sashank Santhanam}, \bibinfo{person}{Behnam
  Hedayatnia}, \bibinfo{person}{Spandana Gella}, \bibinfo{person}{Aishwarya
  Padmakumar}, \bibinfo{person}{Seokhwan Kim}, \bibinfo{person}{Yang Liu},
  {and} \bibinfo{person}{Dilek Hakkani-Tur}.} \bibinfo{year}{2021}\natexlab{}.
\newblock \showarticletitle{Rome was built in 1776: A case study on factual
  correctness in knowledge-grounded response generation}.
\newblock \bibinfo{journal}{\emph{arXiv preprint arXiv:2110.05456}}
  (\bibinfo{year}{2021}).
\newblock


\bibitem[Scao et~al\mbox{.}(2022)]%
        {scao2022bloom}
\bibfield{author}{\bibinfo{person}{Teven~Le Scao}, \bibinfo{person}{Angela
  Fan}, \bibinfo{person}{Christopher Akiki}, \bibinfo{person}{Ellie Pavlick},
  \bibinfo{person}{Suzana Ili{\'c}}, \bibinfo{person}{Daniel Hesslow},
  \bibinfo{person}{Roman Castagn{\'e}}, \bibinfo{person}{Alexandra~Sasha
  Luccioni}, \bibinfo{person}{Fran{\c{c}}ois Yvon}, \bibinfo{person}{Matthias
  Gall{\'e}}, {et~al\mbox{.}}} \bibinfo{year}{2022}\natexlab{}.
\newblock \showarticletitle{Bloom: A 176b-parameter open-access multilingual
  language model}.
\newblock \bibinfo{journal}{\emph{arXiv preprint arXiv:2211.05100}}
  (\bibinfo{year}{2022}).
\newblock


\bibitem[Scialom et~al\mbox{.}(2021)]%
        {scialom2021questeval}
\bibfield{author}{\bibinfo{person}{Thomas Scialom},
  \bibinfo{person}{Paul-Alexis Dray}, \bibinfo{person}{Patrick Gallinari},
  \bibinfo{person}{Sylvain Lamprier}, \bibinfo{person}{Benjamin Piwowarski},
  \bibinfo{person}{Jacopo Staiano}, {and} \bibinfo{person}{Alex Wang}.}
  \bibinfo{year}{2021}\natexlab{}.
\newblock \showarticletitle{Questeval: Summarization asks for fact-based
  evaluation}.
\newblock \bibinfo{journal}{\emph{arXiv preprint arXiv:2103.12693}}
  (\bibinfo{year}{2021}).
\newblock


\bibitem[Shen et~al\mbox{.}(2023)]%
        {shen2023chatgpt}
\bibfield{author}{\bibinfo{person}{Xinyue Shen}, \bibinfo{person}{Zeyuan Chen},
  \bibinfo{person}{Michael Backes}, {and} \bibinfo{person}{Yang Zhang}.}
  \bibinfo{year}{2023}\natexlab{}.
\newblock \showarticletitle{In ChatGPT We Trust? Measuring and Characterizing
  the Reliability of ChatGPT}.
\newblock \bibinfo{journal}{\emph{arXiv preprint arXiv:2304.08979}}
  (\bibinfo{year}{2023}).
\newblock


\bibitem[Shuster et~al\mbox{.}(2021)]%
        {shuster2021retrieval}
\bibfield{author}{\bibinfo{person}{Kurt Shuster}, \bibinfo{person}{Spencer
  Poff}, \bibinfo{person}{Moya Chen}, \bibinfo{person}{Douwe Kiela}, {and}
  \bibinfo{person}{Jason Weston}.} \bibinfo{year}{2021}\natexlab{}.
\newblock \showarticletitle{Retrieval augmentation reduces hallucination in
  conversation}.
\newblock \bibinfo{journal}{\emph{arXiv preprint arXiv:2104.07567}}
  (\bibinfo{year}{2021}).
\newblock


\bibitem[Su et~al\mbox{.}(2020)]%
        {su2020diversifying}
\bibfield{author}{\bibinfo{person}{Hui Su}, \bibinfo{person}{Xiaoyu Shen},
  \bibinfo{person}{Sanqiang Zhao}, \bibinfo{person}{Xiao Zhou},
  \bibinfo{person}{Pengwei Hu}, \bibinfo{person}{Randy Zhong},
  \bibinfo{person}{Cheng Niu}, {and} \bibinfo{person}{Jie Zhou}.}
  \bibinfo{year}{2020}\natexlab{}.
\newblock \showarticletitle{Diversifying dialogue generation with
  non-conversational text}.
\newblock \bibinfo{journal}{\emph{arXiv preprint arXiv:2005.04346}}
  (\bibinfo{year}{2020}).
\newblock


\bibitem[Tian et~al\mbox{.}(2019)]%
        {tian2019sticking}
\bibfield{author}{\bibinfo{person}{Ran Tian}, \bibinfo{person}{Shashi Narayan},
  \bibinfo{person}{Thibault Sellam}, {and} \bibinfo{person}{Ankur~P Parikh}.}
  \bibinfo{year}{2019}\natexlab{}.
\newblock \showarticletitle{Sticking to the facts: Confident decoding for
  faithful data-to-text generation}.
\newblock \bibinfo{journal}{\emph{arXiv preprint arXiv:1910.08684}}
  (\bibinfo{year}{2019}).
\newblock


\bibitem[Touvron et~al\mbox{.}(2023)]%
        {touvron2023llama}
\bibfield{author}{\bibinfo{person}{Hugo Touvron}, \bibinfo{person}{Thibaut
  Lavril}, \bibinfo{person}{Gautier Izacard}, \bibinfo{person}{Xavier
  Martinet}, \bibinfo{person}{Marie-Anne Lachaux},
  \bibinfo{person}{Timoth{\'e}e Lacroix}, \bibinfo{person}{Baptiste
  Rozi{\`e}re}, \bibinfo{person}{Naman Goyal}, \bibinfo{person}{Eric Hambro},
  \bibinfo{person}{Faisal Azhar}, {et~al\mbox{.}}}
  \bibinfo{year}{2023}\natexlab{}.
\newblock \showarticletitle{Llama: Open and efficient foundation language
  models}.
\newblock \bibinfo{journal}{\emph{arXiv preprint arXiv:2302.13971}}
  (\bibinfo{year}{2023}).
\newblock


\bibitem[Trischler et~al\mbox{.}(2016)]%
        {trischler2016newsqa}
\bibfield{author}{\bibinfo{person}{Adam Trischler}, \bibinfo{person}{Tong
  Wang}, \bibinfo{person}{Xingdi Yuan}, \bibinfo{person}{Justin Harris},
  \bibinfo{person}{Alessandro Sordoni}, \bibinfo{person}{Philip Bachman}, {and}
  \bibinfo{person}{Kaheer Suleman}.} \bibinfo{year}{2016}\natexlab{}.
\newblock \showarticletitle{Newsqa: A machine comprehension dataset}.
\newblock \bibinfo{journal}{\emph{arXiv preprint arXiv:1611.09830}}
  (\bibinfo{year}{2016}).
\newblock


\bibitem[Wang et~al\mbox{.}(2020a)]%
        {wang2020asking}
\bibfield{author}{\bibinfo{person}{Alex Wang}, \bibinfo{person}{Kyunghyun Cho},
  {and} \bibinfo{person}{Mike Lewis}.} \bibinfo{year}{2020}\natexlab{a}.
\newblock \showarticletitle{Asking and answering questions to evaluate the
  factual consistency of summaries}.
\newblock \bibinfo{journal}{\emph{arXiv preprint arXiv:2004.04228}}
  (\bibinfo{year}{2020}).
\newblock


\bibitem[Wang and Komatsuzaki(2021)]%
        {gpt-j}
\bibfield{author}{\bibinfo{person}{Ben Wang} {and} \bibinfo{person}{Aran
  Komatsuzaki}.} \bibinfo{year}{2021}\natexlab{}.
\newblock \bibinfo{title}{GPT-J-6B: A 6 Billion Parameter Autoregressive
  Language Model}.
\newblock
  \bibinfo{howpublished}{\url{https://github.com/kingoflolz/mesh-transformer-jax}}.
\newblock
\urldef\tempurl%
\url{https://github.com/kingoflolz/mesh-transformer-jax}
\showURL{%
\tempurl}


\bibitem[Wang et~al\mbox{.}(2020b)]%
        {wang2020towards}
\bibfield{author}{\bibinfo{person}{Zhenyi Wang}, \bibinfo{person}{Xiaoyang
  Wang}, \bibinfo{person}{Bang An}, \bibinfo{person}{Dong Yu}, {and}
  \bibinfo{person}{Changyou Chen}.} \bibinfo{year}{2020}\natexlab{b}.
\newblock \showarticletitle{Towards faithful neural table-to-text generation
  with content-matching constraints}.
\newblock \bibinfo{journal}{\emph{arXiv preprint arXiv:2005.00969}}
  (\bibinfo{year}{2020}).
\newblock


\bibitem[Williams et~al\mbox{.}(2017)]%
        {williams2017broad}
\bibfield{author}{\bibinfo{person}{Adina Williams}, \bibinfo{person}{Nikita
  Nangia}, {and} \bibinfo{person}{Samuel~R Bowman}.}
  \bibinfo{year}{2017}\natexlab{}.
\newblock \showarticletitle{A broad-coverage challenge corpus for sentence
  understanding through inference}.
\newblock \bibinfo{journal}{\emph{arXiv preprint arXiv:1704.05426}}
  (\bibinfo{year}{2017}).
\newblock


\bibitem[Xia et~al\mbox{.}(2024)]%
        {xia2024aicodereval}
\bibfield{author}{\bibinfo{person}{Yinghui Xia}, \bibinfo{person}{Yuyan Chen},
  \bibinfo{person}{Tianyu Shi}, \bibinfo{person}{Jun Wang}, {and}
  \bibinfo{person}{Jinsong Yang}.} \bibinfo{year}{2024}\natexlab{}.
\newblock \showarticletitle{AICoderEval: Improving AI Domain Code Generation of
  Large Language Models}.
\newblock \bibinfo{journal}{\emph{arXiv preprint arXiv:2406.04712}}
  (\bibinfo{year}{2024}).
\newblock


\bibitem[Yan and Xu(2023)]%
        {yan2023refining}
\bibfield{author}{\bibinfo{person}{Tianqiang Yan} {and}
  \bibinfo{person}{Tiansheng Xu}.} \bibinfo{year}{2023}\natexlab{}.
\newblock \showarticletitle{Refining the Responses of LLMs by Themselves}.
\newblock \bibinfo{journal}{\emph{arXiv preprint arXiv:2305.04039}}
  (\bibinfo{year}{2023}).
\newblock


\bibitem[Yang et~al\mbox{.}(2021)]%
        {yang2021amqan}
\bibfield{author}{\bibinfo{person}{Haitian Yang}, \bibinfo{person}{Weiqing
  Huang}, \bibinfo{person}{Xuan Zhao}, \bibinfo{person}{Yan Wang},
  \bibinfo{person}{Yuyan Chen}, \bibinfo{person}{Bin Lv}, \bibinfo{person}{Rui
  Mao}, {and} \bibinfo{person}{Ning Li}.} \bibinfo{year}{2021}\natexlab{}.
\newblock \showarticletitle{Amqan: Adaptive multi-attention question-answer
  networks for answer selection}. In \bibinfo{booktitle}{\emph{Machine Learning
  and Knowledge Discovery in Databases: European Conference, ECML PKDD 2020,
  Ghent, Belgium, September 14--18, 2020, Proceedings, Part III}}. Springer,
  \bibinfo{pages}{584--599}.
\newblock


\bibitem[Yang et~al\mbox{.}(2018)]%
        {yang2018hotpotqa}
\bibfield{author}{\bibinfo{person}{Zhilin Yang}, \bibinfo{person}{Peng Qi},
  \bibinfo{person}{Saizheng Zhang}, \bibinfo{person}{Yoshua Bengio},
  \bibinfo{person}{William~W Cohen}, \bibinfo{person}{Ruslan Salakhutdinov},
  {and} \bibinfo{person}{Christopher~D Manning}.}
  \bibinfo{year}{2018}\natexlab{}.
\newblock \showarticletitle{HotpotQA: A dataset for diverse, explainable
  multi-hop question answering}.
\newblock \bibinfo{journal}{\emph{arXiv preprint arXiv:1809.09600}}
  (\bibinfo{year}{2018}).
\newblock


\bibitem[Ye et~al\mbox{.}(2023)]%
        {ye2023assessing}
\bibfield{author}{\bibinfo{person}{Wentao Ye}, \bibinfo{person}{Mingfeng Ou},
  \bibinfo{person}{Tianyi Li}, \bibinfo{person}{Xuetao Ma},
  \bibinfo{person}{Yifan Yanggong}, \bibinfo{person}{Sai Wu},
  \bibinfo{person}{Jie Fu}, \bibinfo{person}{Gang Chen}, \bibinfo{person}{Junbo
  Zhao}, {et~al\mbox{.}}} \bibinfo{year}{2023}\natexlab{}.
\newblock \showarticletitle{Assessing Hidden Risks of LLMs: An Empirical Study
  on Robustness, Consistency, and Credibility}.
\newblock \bibinfo{journal}{\emph{arXiv preprint arXiv:2305.10235}}
  (\bibinfo{year}{2023}).
\newblock


\bibitem[Zhang et~al\mbox{.}(2019)]%
        {zhang2019bertscore}
\bibfield{author}{\bibinfo{person}{Tianyi Zhang}, \bibinfo{person}{Varsha
  Kishore}, \bibinfo{person}{Felix Wu}, \bibinfo{person}{Kilian~Q Weinberger},
  {and} \bibinfo{person}{Yoav Artzi}.} \bibinfo{year}{2019}\natexlab{}.
\newblock \showarticletitle{Bertscore: Evaluating text generation with bert}.
\newblock \bibinfo{journal}{\emph{arXiv preprint arXiv:1904.09675}}
  (\bibinfo{year}{2019}).
\newblock


\bibitem[Zhao et~al\mbox{.}(2023)]%
        {zhao2023survey}
\bibfield{author}{\bibinfo{person}{Wayne~Xin Zhao}, \bibinfo{person}{Kun Zhou},
  \bibinfo{person}{Junyi Li}, \bibinfo{person}{Tianyi Tang},
  \bibinfo{person}{Xiaolei Wang}, \bibinfo{person}{Yupeng Hou},
  \bibinfo{person}{Yingqian Min}, \bibinfo{person}{Beichen Zhang},
  \bibinfo{person}{Junjie Zhang}, \bibinfo{person}{Zican Dong},
  {et~al\mbox{.}}} \bibinfo{year}{2023}\natexlab{}.
\newblock \showarticletitle{A survey of large language models}.
\newblock \bibinfo{journal}{\emph{arXiv preprint arXiv:2303.18223}}
  (\bibinfo{year}{2023}).
\newblock


\bibitem[Zhou et~al\mbox{.}(2020)]%
        {zhou2020detecting}
\bibfield{author}{\bibinfo{person}{Chunting Zhou}, \bibinfo{person}{Graham
  Neubig}, \bibinfo{person}{Jiatao Gu}, \bibinfo{person}{Mona Diab},
  \bibinfo{person}{Paco Guzman}, \bibinfo{person}{Luke Zettlemoyer}, {and}
  \bibinfo{person}{Marjan Ghazvininejad}.} \bibinfo{year}{2020}\natexlab{}.
\newblock \showarticletitle{Detecting hallucinated content in conditional
  neural sequence generation}.
\newblock \bibinfo{journal}{\emph{arXiv preprint arXiv:2011.02593}}
  (\bibinfo{year}{2020}).
\newblock


\bibitem[Zong et~al\mbox{.}(2024)]%
        {zong2024proswitch}
\bibfield{author}{\bibinfo{person}{Chang Zong}, \bibinfo{person}{Yuyan Chen},
  \bibinfo{person}{Weiming Lu}, \bibinfo{person}{Jian Shao}, {and}
  \bibinfo{person}{Yueting Zhuang}.} \bibinfo{year}{2024}\natexlab{}.
\newblock \showarticletitle{ProSwitch: Knowledge-Guided Language Model
  Fine-Tuning to Generate Professional and Non-Professional Styled Text}.
\newblock \bibinfo{journal}{\emph{arXiv preprint arXiv:2403.09131}}
  (\bibinfo{year}{2024}).
\newblock


\end{thebibliography}

\end{document}